\documentclass[journal]{IEEEtran}
\usepackage{url}
\usepackage{cite}
\usepackage{amsmath,amssymb,amsfonts}
\usepackage{algorithmic}
\usepackage{graphicx}
\usepackage{textcomp}
\usepackage{xcolor}
\usepackage{subfig}
\usepackage{mathtools}

\def\BibTeX{{\rm B\kern-.05em{\sc i\kern-.025em b}\kern-.08em
    T\kern-.1667em\lower.7ex\hbox{E}\kern-.125emX}}
\def\chartheight{4cm}
\usepackage{url}
\usepackage{todonotes}

\usepackage{pgfplots}
\usepackage{flushend}
\usepackage{color}
\usepackage{capt-of}
\usetikzlibrary{shadows,patterns,shapes,arrows,decorations.pathmorphing,backgrounds,positioning,fit,plotmarks,calc,spy}
\usepgfplotslibrary{patchplots}
\usetikzlibrary{pgfplots.statistics}
\usetikzlibrary{snakes}

\pgfplotsset{ /pgfplots/ybar legend/.style={
		/pgfplots/legend image code/.code={
			\draw[##1,/tikz/.cd,bar width=3pt,yshift=-0.2em,bar shift=0pt]
			plot coordinates {(0cm,0.8em)};},
	},
}



\usepackage{float}
\usepackage{mdwlist}
\usepackage[linesnumbered,ruled,vlined]{algorithm2e}
\SetAlFnt{\scriptsize}
\usepackage{cases}
\usepackage{graphicx}
\usepackage{multirow}
\usepackage{blindtext}
\usepackage{etoolbox}
\makeatletter
\patchcmd{\maketitle}{\@copyrightspace}{}{}{}
\makeatother
\usepackage{enumitem}
\usepackage{nowidow}
\setitemize{leftmargin=*}
    
\begin{document}

\title{High-Throughput CNN Inference on Embedded ARM big.LITTLE Multi-Core Processors}

\author{ 
	Siqi~Wang, 
	Gayathri~Ananthanarayanan,
	Yifan~Zeng,
	Neeraj~Goel,
	Anuj~Pathania, 
	Tulika~Mitra
	\thanks{ 
		Manuscript received March 14, 2019; accepted September 17, 2019.
		This work was partially funded by Singapore Ministry of Education Academic Research Fund Tier 2 MOE2015-T2-2-088.
		S. Wang, Y. Zeng, A. Pathania and T. Mitra are with the Department of Computer Science, School of Computing, National University of Singapore, SG. 
		E-mail:~((wangsq, yifan122, pathania, tulika)@comp.nus.edu.sg). 
		G. Ananthanarayanan is with the Department of Computer Science and Engineering, Indian Institute of Technology Dharwad, Karnataka, IN E-mail:~(gayathri@iitdh.ac.in). 
		N. Goel is with the Department of Computer Science and Engineering, Indian Institute of Technology Ropar, IN. E-mail:~(neeraj@iitrpr.ac.in). 
		\protect(\em Corresponding author: Tulika Mitra)\protect\\
	}
}
\maketitle

\begin{abstract}
IoT Edge intelligence requires Convolutional Neural Network~(CNN) inference to take place in the edge devices itself. {\em ARM big.LITTLE} architecture is at the heart of prevalent commercial edge devices. It comprises of single-ISA heterogeneous cores grouped into multiple homogeneous clusters that enable power and performance trade-offs. All cores are expected to be simultaneously employed in inference to attain maximal throughput. However, high communication overhead involved in parallelization of computations from convolution kernels across clusters is detrimental to throughput. We present an alternative framework called {\em Pipe-it} that employs pipelined design to split convolutional layers across clusters while limiting parallelization of their respective kernels to the assigned cluster. We develop a performance-prediction model that utilizes only the convolutional layer descriptors to predict the execution time of each layer individually on all permitted core configurations (type and count). {\em Pipe-it} then exploits the predictions to create a balanced pipeline using an efficient design space exploration algorithm. \textit{Pipe-it} on average results in a 39\% higher throughput than the highest antecedent throughput.
\end{abstract}
	
\begin{IEEEkeywords}
		Heterogeneous Multi-Core, Asymmetric Multi-Core, Edge Inference, CNN Performance-Prediction
\end{IEEEkeywords}

\section{Introduction}\label{sec:introduction}

\IEEEPARstart{C}{onvolutional} Neural Network~(CNN) inference on edge devices has become quintessential for enriched user experience. Continuous vision tasks that use inference to extract high-level semantic information from real-time video streams are paramount in numerous edge application domains such as Advanced Driver-Assistance Systems~(ADAS), Virtual Reality~(VR), and Augmented Reality~(AR)~\cite{krizhevsky2012imagenet}. Inference-driven applications project unprecedented computational requirements onto underlying edge devices~\cite{zhu2018mobile}. Fortunately, there has been tremendous progress to port CNNs to edge devices. Many network models such as {\em MobileNet}~\cite{mobilenet} have been invented specifically for edge to perform high-accuracy classifications with considerably smaller network size. Numerous efficient libraries such as {\em ARM Compute Library~(ARM-CL)}~\cite{arm_cl} and {\em Tencent NCNN}~\cite{ncnn} have been constructed precisely to facilitate efficient CNN implementation for the edge. {\em ARM-CL} is highly optimized for edge-specific {\em ARM} core architectures with inbuilt support for multi-threading and acceleration through {\em ARM NEON} vectorization technology. 

\begin{figure}[!t]
	\centering
	\scriptsize
	\sffamily
	\begin{tikzpicture}[x=12, y=9]
	
	\draw[fill=white, draw=white] (0,12) rectangle +(9,1.5) node[pos=.5,] {{\em Big} Cluster};
	\draw[fill=white, rounded corners] (0,2) rectangle +(9,10.25);
	\draw[fill=white, rounded corners] (.25,9.5) rectangle +(4,2) node[pos=.5,] {Cortex A73};
	\draw[fill=white, rounded corners] (4.75,9.5) rectangle +(4,2) node[pos=.5,] {Cortex A73};	
	\draw[fill=white, rounded corners] (.25,6.5) rectangle +(4,2) node[pos=.5,] {Cortex A73};
	\draw[fill=white, rounded corners] (4.75,6.5) rectangle +(4,2) node[pos=.5,] {Cortex A73};
	\draw[fill=white, rounded corners] (0.25,5) rectangle +(8.5,1) node[pos=.5,] {\tiny SCU};
	\draw[fill=white, rounded corners] (1.5,2.5) rectangle +(6,2) node[pos=.5,] {2\,MB L2 Cache};
	
	\draw[fill=white, draw=white] (10,11.25) rectangle +(8,1.5) node[pos=.5,] {{\em Small} Cluster};
	\draw[fill=white, rounded corners] (10,2) rectangle +(8,9.25);
	\draw[fill=white, rounded corners] (10.25,8.75) rectangle +(3.5,2) node[pos=.5,] {Cortex A53};
	\draw[fill=white, rounded corners] (14.25,8.75) rectangle +(3.5,2) node[pos=.5,] {Cortex A53};
	\draw[fill=white, rounded corners] (10.25,6.25) rectangle +(3.5,2) node[pos=.5,] {Cortex A53};
	\draw[fill=white, rounded corners] (14.25,6.25) rectangle +(3.5,2) node[pos=.5,] {Cortex A53};
	\draw[fill=white, rounded corners] (10.25,5) rectangle +(7.5,1) node[pos=.5,] {\tiny SCU};
	\draw[fill=white, rounded corners] (11,2.5) rectangle +(6,2) node[pos=.5,] {1\,MB L2 Cache};
	
	\draw[fill=white, rounded corners] (0,-1) rectangle +(18,2) node[pos=.5,] {CCI Bus};
	
	\end{tikzpicture}
	\caption{An abstract block diagram of an eight-core {\em ARM big.LITTLE} heterogeneous multi-core within {\em Hi3670} SoC~\cite{hikey970}.}
	\label{fig:Hi3670}
\end{figure}
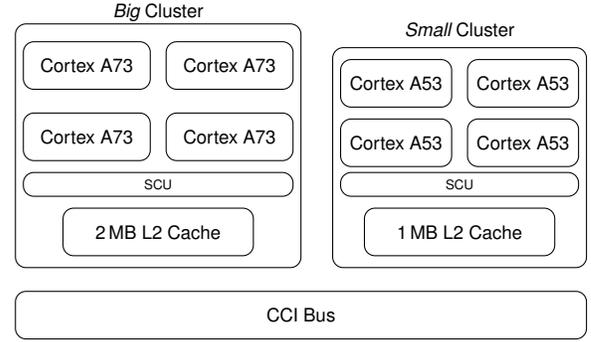

Single-ISA heterogeneous multi-cores comprise of processing cores that have different power-performance-area characteristics but share the same Instruction Set Architecture~(ISA)~\cite{kumar2003single}. Facebook~\cite{wu2019machine} in 2019 reports that about half of the mobile SoCs in the market adopts such architecture with two CPU clusters: a high-performance cluster and an energy-efficient cluster. This heterogeneous configuration provides higher parallel processing potential than homogeneous multi-cores within given power and area budget provided all cores can be simultaneously employed productively~\cite{muthukaruppan2014price,muthukaruppan2013hierarchical}. 
Figure~\ref{fig:Hi3670} shows an abstract block diagram for the eight-core state-of-the-art {\em ARM big.LITTLE} heterogeneous multi-core in {\em Hi3670} System on Chip~(SoC) designed for edge devices. 
{\em Hi3670} groups together four high-performance {\em Big Cortex A73} cores and four low-performance {\em Small Cortex A53} cores into two clusters alongside L2 caches of size 2\,MB and 1\,MB, respectively. 
Two clusters are kept fully cache-coherent via bus-based Cache Coherent Interconnect~(CCI) using snooping broadcast protocol. 
Cores within a cluster are kept coherent using bus-based Snoop Control Unit~(SCU). 
This raw computational power provided by heterogeneous multi-core makes CNN inference on edge device feasible.

\begin{figure*}[!t]
	\centering
	\includegraphics[width=0.95\textwidth]{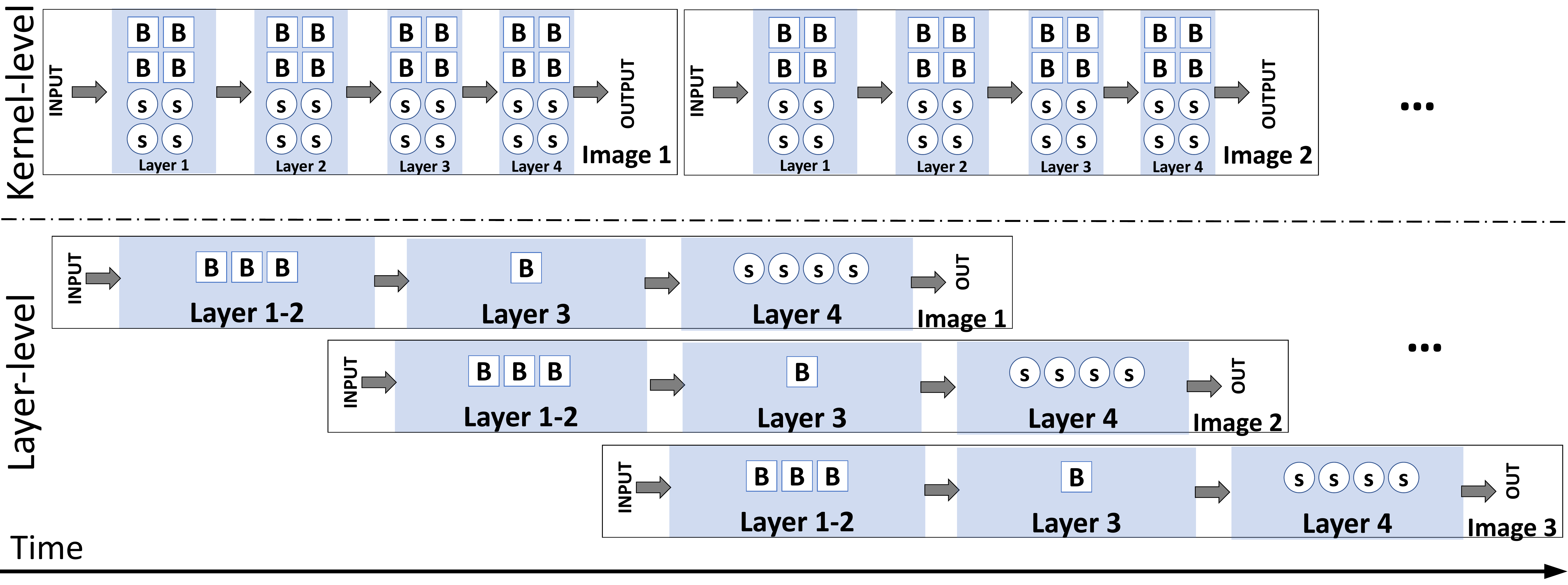}
	\caption{Visualization of the default {\em Kernel-level} and the proposed {\em Layer-level} splitting with a three-stage pipeline {\em (B3-B1-s4)} on heterogeneous multi-core with four {\em Big (B)} cores and four {\em Small (s)} cores for a representational four-layer CNN.}
	\label{fig:split_techniques}
\end{figure*}

Dedicated accelerators such as GPUs and dedicated IP cores have been proven to be more efficient than CPU for inference. However, their applicability is constrained by the extreme diversity of accelerators and lack of easy programming support. CPU remains the platform of choice for running ML workloads being the most common denominator with high availability in mobile and embedded platforms~\cite{wu2019machine,liu2018optimizing,tang2018scheduling,znn2016}.
In addition, low-cost edge devices may not contain dedicated accelerators, and the performance gap between CPU and GPU is small, making CPU the favourable choice for ML workloads. On the other hand, CNNs are more commonly used as a building block to construct more complex systems. For applications ranging from smart classroom~\cite{smart-classroom} with person and text recognition, to autonomous drones~\cite{drone} with path planning, object classification and obstacle avoidance, multiple independent inference sub-tasks are performed concurrently. Such applications require all the available resources to run these inference engines in parallel. {\em Therefore, improving inference throughput on \textit{ARM big.LITTLE} like architectures by itself is a critical problem.}

\noindent \textbf {Motivational Example:} 
The layers in a CNN are in a pre-ordained order by design, which is usually in sequential. Their associated convolutional kernels are therefore required to be processed sequentially.
Nevertheless, different images from an image stream can potentially be processed in parallel. Unfortunately, existing state-of-the-art deep learning libraries such as {\em ARM-CL} is designed to process the image stream sequentially one image at a time. The computation of one kernel at a time is then distributed across all cores with the default parallelization strategy we christen {\em Kernel-level} execution. Figure~\ref{fig:split_techniques} (top) visualizes the {\em Kernel-level} strategy for a representative four-layer CNN on a eight-core heterogeneous multi-core. 
Section~\ref{sec:ARM} provides further details on {\em Kernel-level} strategy.  
The {\em Kernel-level} strategy works for intra-cluster processing but fails to scale to inter-cluster processing with multiple clusters.

\begin{figure} [!t]
	\centering
	\scriptsize
	\begin{tikzpicture}
	\pgfplotsset{set layers}
	\begin{axis}[
	width=\columnwidth, height=\chartheight,
	xlabel={Core Configurations},
	ylabel={Throughput [Images/Second]},
	xtick=data,
	xticklabels from table={Data/Default_Split.data}{Configuration},
	xlabel style={yshift=-7pt},
	x tick label style={rotate=45,anchor=east},
	bar width=3pt,
	ymajorgrids=true,
	yminorgrids=true,
	ymin=1,		
	legend columns = 3,	
	legend style={at={(0.5,1.05)}, anchor=south, font=\scriptsize},
	cycle list name=black white,
	]
	
	\addlegendentry{\em AlexNet}  \addplot [black, mark=otimes]table[x expr = \coordindex, y = alexnet, ]{Data/Default_Split.data};
	\addlegendentry{\em GoogLeNet} \addplot [red, mark=square] table[x expr = \coordindex, y = googlenet,]{Data/Default_Split.data};		        
	\addlegendentry{\em MobileNet}  \addplot [blue, mark=triangle] table[x expr = \coordindex, y = mobilenet,]{Data/Default_Split.data};
	\addlegendentry{\em ResNet50}  \addplot [lime, mark=diamond] table[x expr = \coordindex, y = resnet,]{Data/Default_Split.data};
	\addlegendentry{\em SqueezeNet}  \addplot [brown, mark=star] table[x expr = \coordindex, y = squeezenet, ]{Data/Default_Split.data};
	
	\end{axis}
	\end{tikzpicture}
	\caption {Throughput of different CNNs with a different number of heterogeneous cores (B: {\em Big} core, s: {\em Small} core) using the default {\em Kernel-level} strategy.}
	\label{fig:CNN_Default_Split}
\end{figure}
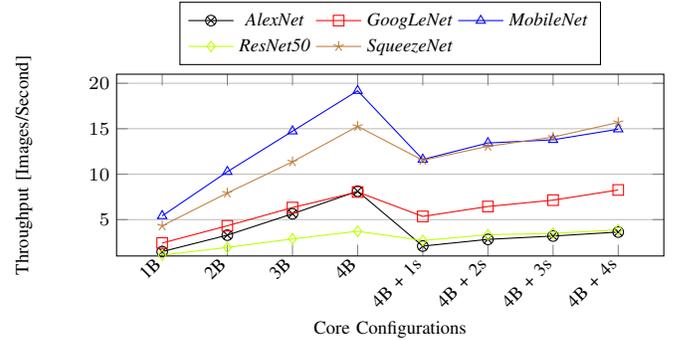

Heterogeneous Multi-Processing~(HMP) allows execution of kernels using both {\em Big} and {\em Small} Cores simultaneously. Figure~\ref{fig:CNN_Default_Split} shows the change in throughput (measured in images per second) of several CNNs with the increase in the number of heterogeneous cores used with {\em Kernel-level} strategy. Throughput increases as we add more  {\em Big} cores but drops sharply on the addition of {\em Small} cores from another cluster for HMP. Inter-cluster communication overhead involved in the use of HMP explains the drop. No HMP configuration surpasses the performance of configuration with four {\em Big} cores. Therefore, \textit{Figure~\ref{fig:CNN_Default_Split} empirically shows that we cannot improve throughput on heterogeneous multi-cores with default Kernel-Level strategy alone.} This limitation originates from the design of  {\em Kernel-level} strategy and not from the quality of its implementation.

There are multiple convolutional layers of different dimensions within a CNN  that project different resource requirements. Therefore, it is possible to create a processing pipeline with stages composed of only homogeneous cores that still splits CNN processing over different heterogeneous clusters. Let notation \textit{\{core\_type\}\{core\_count\}}  denote the core configuration of a pipeline stage. Figure~\ref{fig:split_techniques} shows a three-stage pipeline created to process incoming images in a stream using the {\em Layer-level} strategy. Three {\em Big} cores (\textit{B3}) construct first pipeline stage processing Layers 1 and 2. Remaining one {\em Big} core (\textit{B1}) constructs second stage processing Layer 3. Four {\em Small} cores (\textit{s4}) construct third pipeline stage processing Layer 4. This pipeline constructively uses all eight heterogeneous cores in execution by processing multiple images in parallel. Generally, initial layers operating on bigger inputs requires more computational power and memory compared to deeper layers. Therefore, it is intuitive to map initial convolutional layers to more powerful {\em Big} cluster and deeper layers to less powerful {\em Small} cluster. However, the design space of mapping layers to core clusters increases exponentially with the increase in the number of layers.

\noindent \textbf{Our Novel Contributions}: 
We propose a framework called {\em Pipe-it} that partitions CNN layers across heterogeneous cores to improve throughput. {\em Pipe-it} creates a processing pipeline by splitting layers among heterogeneous core clusters, wherein a given set of homogeneous core(s) always process kernels from a fixed set of layers. Different pipeline stages (and cores within) are responsible for concurrently processing different layers corresponding to consecutive images in a stream. The pipelined execution improves throughput by employing all on-chip memory and processing resources of heterogeneous multi-core more effectively than the default approach of splitting individual kernels across all heterogeneous cores. 

{\em Pipe-it}  includes an analytical performance model that predicts the performance of convolutional layer on different core configurations (type and count) from its network structure description. 
Its Design Space Exploration~(DSE) algorithm then uses the predicted performance to locate the best fitting pipeline configuration and respective layer allocation.
On average, we get 39\% improvement in throughput from entire heterogeneous multi-core compared to using only its high-performance homogeneous {\em Big} cluster.

\section{Background} \label{sec:ARM}

\begin{figure} [!t]
	\centering
	\scriptsize
	\begin{tikzpicture}
	
	\pgfplotsset{set layers}
	
	\begin{axis}[ 
	width=\columnwidth, height=\chartheight,  
	ylabel={\scriptsize Throughput [Images/Sec] }, 
	bar width=8pt, 
	xtick=data,
	ybar, 
	xticklabels from table={Data/SoA.data}{CNN}, x tick label style={rotate=45,anchor=east, font=\scriptsize}, 
	ymajorgrids=true, 
	yminorgrids=true, 
	ymin=0, 
	legend columns=3,
	legend style={at={(0.5,1.05)}, anchor=south, cells={anchor=west}, font=\footnotesize, inner sep=0.2pt},
	legend entries={{\em ARM-CL~\cite{arm_cl}},{\em NCNN~\cite{ncnn}}, {\em TVM~\cite{chen2018tvm}}},
	]

	\addplot[ postaction={ pattern=horizontal lines}] table[ x expr = \coordindex, y = ARM-CL, ]{Data/SoA.data};
	\addplot[ postaction={ pattern=crosshatch}] table[ x expr = \coordindex, y = NCNN, ]{Data/SoA.data};
	\addplot[ postaction={ pattern=dots}] table[ x expr = \coordindex, y = TVM, ]{Data/SoA.data};

	\end{axis}
	\end{tikzpicture}
	*{\em TVM} results are generated with \textit{NNVM-TVM} framework with a pre-trained model from \textit{mxnet.gluon.mode\_zoo.vision} model set~\cite{mxnet_zoo}, wherein \textit{GoogLeNet} is not included.
	\caption{Throughput of different CNN models on {\em Big} cluster when implemented in different deep learning frameworks.}
	\label {fig:SOA_ARMCL}
\end{figure}

{\em ARM Compute Library~(ARM-CL)}~\cite{arm_cl} is a state-of-the-art framework for implementing CNNs on {\em ARM} architectures. 
Figure~\ref{fig:SOA_ARMCL} shows the throughput of CNN inference implemented with {\em ARM-CL (version 18.05)}, {\em Tencent NCNN}~\cite{ncnn}, and {\em TVM}~\cite{chen2018tvm} frameworks running on {\em Big} cluster using multi-threading. 
Both {\em ARM-CL} and {\em Tencent NCNN} support acceleration through {\em ARM NEON} vectorization and provides  {\em NEON} assembly implementation for most computationally intensive convolution kernels of CNN. 
These two frameworks present similar performance and outperform {\em TVM} implementation without {\em NEON} acceleration. 
However, {\em Tencent NCNN} is not as well maintained or supported as {\em ARM-CL}. 
Therefore, we use {\em ARM-CL} as the foundational framework in this work.

\begin{table}[!t]
	{	\centering
		\caption{Structure of different CNN models and the corresponding major layer~(node) counts in their default {\em ARM-CL} implementations.}
		\resizebox{\columnwidth}{!}{
			\begin{tabular}{l|l|r}
				\hline
				\multicolumn{1}{c|}{CNN} & \multicolumn{1}{c|}{Major Layers/Modules} & \multicolumn{1}{c}{\begin{tabular}[c]{@{}c@{}}{\em ARM-CL} \\ Major/(Total  \\ Node Count) \end{tabular}} \\ \hline\hline
				{\em AlexNet}~\cite{krizhevsky2012imagenet} & 5 Conv + 3 FC & 11* / (21) \\ \hline
				{\em GoogLeNet}~\cite{googlenet} & \begin{tabular}[l]{@{}l@{}} 3 Conv + 9 Inception Modules \\ (6 Conv Each) + 1 FC \end{tabular} & 58 / (132) \\ \hline
				{\em MobileNet}~\cite{mobilenet} & 14 Conv + 13 Conv DW + 1 FC & 28 / (58) \\ \hline
				{\em ResNet50}~\cite{resnet} & \begin{tabular}[l]{@{}l@{}} 1 Conv + 4 Residual Blocks \\ (52 Conv in Total) + 1 FC \end{tabular} & 54 / (146)\\ \hline
				{\em SqueezeNet}~\cite{squeezenet} & \begin{tabular}[l]{@{}l@{}} 2 Conv + 8 Fire Module \\ (3 Conv Each)\end{tabular} & 26 / (58) \\ \hline
			\end{tabular}
		}
		\label{tab:cnn_structure}
	}\\
	Conv: Convolutional Layers; FC: Fully-connected Layers; Conv DW: Depthwise Convolutional Layers. *Three convolutional layers are implemented as two nodes each for {\em AlexNet}.
\end{table}
{\em ARM-CL} is a collection of functions commonly used in machine learning. 
The functions are infused with hardware-specific optimizations for superior performance on {\em ARM} architectures.
\textit{Graph} API accompanying {\em ARM-CL} facilitates the creation of complex networks. The network is written with dedicated API as a graph by the user at the frontend. The execution is automatically handled at the backend. Graph implements the layers as nodes that are connected to other nodes in the CNN sequence as defined by the user. Table~\ref{tab:cnn_structure} summarizes the architecture of several popular CNNs and their respective implementations in {\em ARM-CL}. We count weighted layers~(convolutional or fully-connected) as major layers because they are, in general, most computationally expensive part of CNNs.

Inside each node, the workload is represented as a series of compute kernels. 
Runtime scheduler sequentially dispatches the kernels in q node and engages respective processing unit during execution. 
{\em ARM-CL} implements a convolution node with {\em NEON} acceleration using {\em im2col}~(Image to Column) and GEMM~(GEneral Matrix Multiplication) kernels. In addition, the parallel nature of the kernels allows their computations to be distributed across multiple cores. This node-level parallelization is implemented in the form of a thread pool that spawns several new threads and distributes the computation of a kernel among them before the scheduler dispatches them for execution.

We extended the default {\em ARM-CL} CNN implementations to execute multiple graphs in parallel. 
The implementation allows the same network to be applied to multiple images concurrently. 
All graphs share the same copy of read-only parameters (weights and biases) and each graph contains its unique copy of the image CNN needs to classify as we assume images in a stream to be independent. 
We modify the scheduler to run under a one-thread-per-core model with minimal migrations using thread pinning for faster and predictable execution. 

\section{Co-Execution at different levels}

\subsection{Kernel-Level Splitting}~\label{Sec:Kernel-Level}
We can explore parallelism inherent in kernels by exploiting {\em ARM-CL} thread pool implementation to engage all cores. 
While the parallelization of a kernel across homogeneous cores within a cluster gives performance benefits, further parallelization across heterogeneous clusters does not improve throughput as shown in Figure \ref{fig:CNN_Default_Split}. 
Authors in~\cite{Damsche2018} make a similar observation for kernel-level splitting in the context of CPU-GPU co-execution. 

Using multiple cores within the same cluster for processing increases parallel L2 accesses per unit time. 
Cluster's SCU successfully handles the increased accesses without being overwhelmed and thereby improves performance.
However, when additional cores from another cluster are engaged, the working set gets split between the L2 caches of two clusters.  
Some conflict misses that occur on one cluster now get served by L2 cache of another cluster using CCI increasing average on-chip L2 access latency.
Additional L2 cache decreases the number of capacity misses going to main memory. 
However, the decrease cannot compensate for the increased latency of conflict misses.

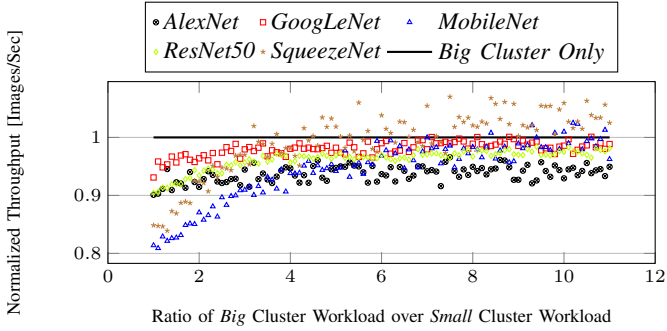
\begin{figure}[!t]
	\centering
	\scriptsize
	\begin{tikzpicture}
	\begin{axis}[
	width=\columnwidth, height=\chartheight,
	xlabel={Ratio of {\em Big} Cluster Workload over {\em Small} Cluster Workload},
	ylabel={Normalized Throughput [Images/Sec]},
	legend style={at={(0.5,1.05)}, anchor=south, cells={anchor=west}, font=\small, },
	legend columns=3,
	ymajorgrids=true,
	]
	
	\addlegendentry{{\em AlexNet}}
	\addplot[
	only marks, mark=otimes, mark size =  1,
	]
	table[x=Factor,y=alexnet,meta=Label]
	{Data/Workload_Split.data};
	
	\addlegendentry{{\em GoogLeNet}}
	\addplot[
	only marks, mark=square, mark size =  1, red,
	]
	table[x=Factor,y=googlenet,meta=Label]
	{Data/Workload_Split.data};
	
	\addlegendentry{{\em MobileNet}}
	\addplot[
	only marks, mark=triangle, mark size =  1, blue,
	]
	table[x=Factor,y=mobilenet,meta=Label]
	{Data/Workload_Split.data};
	
	\addlegendentry{{\em ResNet50}}
	\addplot[
	only marks, mark=diamond, mark size =  1, lime,
	]
	table[x=Factor,y=resnet50,meta=Label]
	{Data/Workload_Split.data};
	
	\addlegendentry{{\em SqueezeNet}}
	\addplot[
	only marks, mark=star, mark size =  1, brown,
	]
	table[x=Factor,y=squeezenet,meta=Label]
	{Data/Workload_Split.data};
	
	\addlegendentry{{\em Big Cluster Only}}
	\addplot[mark=none, solid, thick] coordinates {(1,1) (11,1)};
	
	\end{axis}
	\end{tikzpicture}
	\caption{Throughput of CNN models with disproportionate kernel-level workload split between {\em Big} and {\em Small} cluster normalized against throughput achieved using {\em Big} cluster only.}
	\label{fig:Workload_Split}
\end{figure}

Figure \ref{fig:CNN_Default_Split} shows the throughput obtained by splitting the computational workload from kernel equally among all threads. 
However, distributing workload disproportionately does not improve throughput significantly either.
Figure~\ref{fig:Workload_Split} shows through exhaustive search that no ratio of workload split between {\em Big} and {\em Small} clusters results in statistically significant higher throughput for most CNNs than when kernels run exclusively only on {\em Big} cluster. 
Exhaustive search indicates we must give little or no share of computational work to {\em Small} cluster for optimal execution.

\subsection{Layer-Level Splitting}

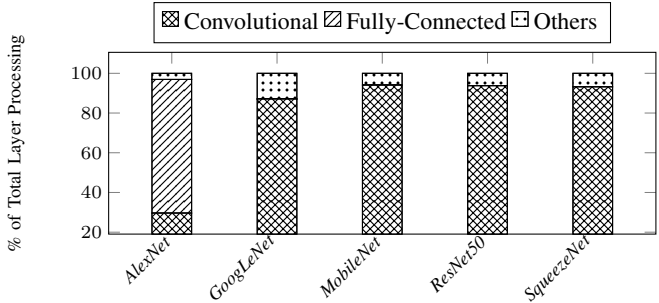
\begin{figure} [!t]
	\centering
	\scriptsize
	\begin{tikzpicture}
	\begin{axis}[
	ybar stacked,
	bar width=15pt,
	width=\columnwidth, height=\chartheight,
	legend columns=3,
	enlargelimits=0.15,
	legend style={at={(0.5,1.05)}, anchor=south, cells={anchor=west}, font=\small, },
	ylabel={\% of Total Layer Processing},
	symbolic x coords={{\em AlexNet}, {\em GoogLeNet}, {\em MobileNet}, {\em ResNet50}, 
		{\em SqueezeNet}},
	xtick=data,
	x tick label style={rotate=45,anchor=east},
	]
	
	\addplot+[ybar, black, fill=white, postaction={ pattern=crosshatch}] plot coordinates {({\em AlexNet},29.6) ({\em GoogLeNet},87.06) ({\em MobileNet},94.06) ({\em ResNet50},93.75) ({\em SqueezeNet},93.18)}; 
	
	\addplot+[ybar, black, fill=white, postaction={ pattern=north east lines}] plot coordinates {({\em AlexNet},67.3) ({\em GoogLeNet},0.01) ({\em MobileNet},0) ({\em ResNet50},0) ({\em SqueezeNet},0)}; 
	
	\addplot+[ybar, black, fill=white, postaction={ pattern=dots}] plot coordinates {({\em AlexNet},3.1) ({\em GoogLeNet},12.93) ({\em MobileNet},5.94) ({\em ResNet50},6.25) ({\em SqueezeNet},6.82)}; 

	\legend{ Convolutional, Fully-Connected, Others, }
	\end{axis}
	\end{tikzpicture}

	\caption{The breakdown of CNN processing time between different layer types.}
	\label {fig:Layer_Split}
	
\end{figure}

Image classification CNNs are made up of multiple layers, which process images sequentially.
Figure~\ref{fig:Layer_Split} shows the share of processing time spent on convolutional layers in different CNNs normalized to total forward pass processing time. 
Processing of convolutional layers dominates overall time spent for all networks except in relatively older {\em AlexNet}, wherein fully-connected layers dominate.

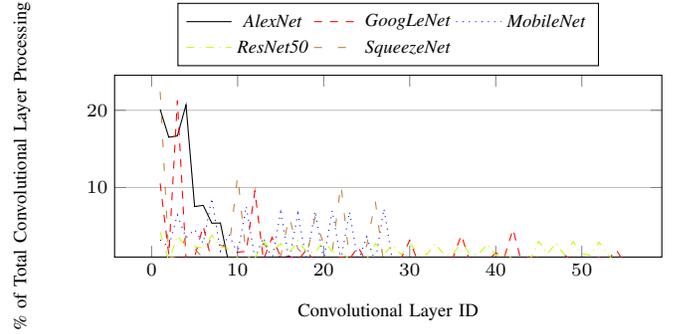
\begin{figure} [!t]
	\centering
	\scriptsize
	\begin{tikzpicture}
	\pgfplotsset{set layers}
	\begin{axis}[
	width=\columnwidth, height=\chartheight,
	xlabel={Convolutional Layer ID},
	ylabel={\% of Total Convolutional Layer Processing},
	bar width=3pt,
	ymajorgrids=true,
	yminorgrids=true,
	ymin=1,		
	legend columns = 3,	
	legend style={at={(0.5,1.05)}, anchor=south, font=\scriptsize},
	cycle list name=black white,
	]
	
	\addlegendentry{\em AlexNet}  \addplot [solid] table[x = Index, y = AlexNet]{Data/Conv_Split.data};
	\addlegendentry{\em GoogLeNet} \addplot [red, dashed]  table[x = Index, y = GoogLeNet,]{Data/Conv_Split.data};		        
	\addlegendentry{\em MobileNet}  \addplot [blue, dotted]  table[x = Index, y = MobileNet,]{Data/Conv_Split.data};
	\addlegendentry{\em ResNet50}  \addplot [lime, dashdotted]  table[x = Index, y = ResNet50,]{Data/Conv_Split.data};
	\addlegendentry{\em SqueezeNet}  \addplot  [brown, loosely dashed] table[x = Index, y = SqueezeNet, ]{Data/Conv_Split.data};
	
	\end{axis}
	\end{tikzpicture}
	
	\caption {Distribution of total convolution processing time among convolutional layers for different networks.}
	\label{fig:ConvolutionProcessingDistribution}
\end{figure}
The convolutional layer at the start of network operates upon the original data of the biggest size (and dimensionality) and produces output data of smaller size due to the application of filters. 
This shrunken output gets passed on to the subsequent convolutional layer as input, which reduces its convolution processing time. 
Figure~\ref{fig:ConvolutionProcessingDistribution} shows that time taken to process convolutional layers generally decreases as we move deeper into a network.

Observations from Figure~\ref{fig:ConvolutionProcessingDistribution}  can help us in creating a load-balanced processing split on a heterogeneous multi-core. 
Its high-performance cores can process more processing-intensive initial layers, while low-performance cores can process less processing-intensive deeper layers.
Kernels from layers can still get split among all homogeneous cores within a cluster using {\em Kernel-level} splitting. 
Kernels from non-convolutional layers are considered part of previous convolutional layers and get processed at the same cluster. 
We do not explore layer-level splitting of CNN at non-convolutional layers.    

{\em Layer-level} splitting between clusters produces a lower number of inter-cluster L2 conflict misses than {\em Kernel-level} splitting as most layers that feed data into each other are processed on the same cluster reducing the load on CCI.   
Furthermore, it also allows for multiple images from a stream to be processed in parallel. 
The {\em Big} cluster can start processing layers from image $Z+1$, while {\em Small} cluster is still processing layers from image $Z$. 
{\em Layer-level} splitting, unlike {\em Kernel-level}, also requires less movement of weight and biases between clusters.  
It processes weights and biases shared between the kernels of different images on the same cluster.
This optimization further reduces the amount of conflict misses between clusters and thereby improves L2 cache usage efficiency.

\begin{table*}[!t]
	\small\centering
	\caption{Description of parameters in chronological order.}
	\resizebox{\textwidth}{!}{
		\begin{tabular}{p{2.65cm} | p{14.9cm}}\hline
			Parameters & Descriptions \\\hline\hline
			$W$ & Workload, number of major layers (convolutional layers, with fully-connected layers for \textit{AlexNet}) in a CNN.\\
			$X, X_1, X_2$& Split-point of workload, number of layers to be allocated to pipeline stages.\\
			$H, H_B, H_s$ & Number of cores in a heterogeneous multi-core architecture. $B$: {\em Big} cores, $s$: {\em Small} cores.\\
			$p, p_B, p_s$ & Number of pipeline stages; number of stages on {\em Big} and {\em Small} clusters.\\
			$C_p$ & Number of different pipeline configurations for a pipeline with $p$ stages.\\
			$D_W$ & Number of design points for a CNN with $W$ major layers on a $H$-core heterogeneous multi-core architecture. \\
			$I_w, I_h, I_d$ & Input image tensor dimensions in width, height and depth. \\
			$F_w, F_h, F_d, Ofm$ & Filter dimensions in width, height, depth and number of output feature maps.\\
			$Pad, S$ & Padding, stride information for convolution. \\
			$N, K, M$ & Dimensions of matrices in convolution converted GEMM. \\
			$\alpha, \beta$ & Regression coefficients. \\
			$ts$ & Tile size for GEMM optimization. \\
			$n_{iter}$ & Number of iterations generated for image tensor with tile size $ts$.\\	
			$iter_t$ & Number of iterations allocated to a thread $t$ in multi-threaded execution.\\	
			$T_{iter}$ & Execution time of a single iteration.\\
			$T_{multi}$ & Execution time of multi-threaded execution.\\
			$P$ =$\{P_1, P_2, ..., P_p\}$ & Representation of a pipeline configuration with $p$ stages\\
			$P_i$ = (type, count) & Representation of the configuration of the $i$-th stage in a pipeline $P$. E.g. $(B, 3)$, also written as $B3$ for convenience.\\
			$L$ =$\{L_1, L_2, ..., L_p\}$ &Corresponding layer allocation for pipeline $P$ with $p$ stages.\\
			$L_i$ =$\{l_j, ..., l_k\}$ & A set of layers in original order allocated to stage $P_i$, also written as $l_{j-k}$ for convenience. \\
			$T, T^{P_i}$ & Time matrix for execution times of a single layer on different core configurations; Time array of execution times of a set of layers with core configuration $P_i$.\\
			$T_{l_j}^{P_i}$, $T_{L_i}^{P_i}$  &  Execution time of layer $l_j$ with pipeline configuration $P_i$; execution time of a pipeline stage $P_i$ with its corresponding layer allocation $L_i$.\\
			$L_{wl}$ & A set of layers as defined in the context (workload).\\
			\hline
	\end{tabular}}\label{tab:params}
\end{table*}

\section {Design Space} \label{design_space}

\subsection{Split Points at Convolutional Layers} \label{subsec:layer_split}

Structure of different convolutional layers can differ significantly from each other within a network. 
Their performance on {\em Big} and {\em Small} clusters with a different number of allocated cores can also be quite different. 
These differences mandate non-trivial decisions on splitting convolutional layers across pipelines stages of {\em Pipe-it}. 

Consider a basic two-stage layer-level split pipeline ({\em B4-s4}) processing a network containing $W$ major layers. 
First $X$ layers are processed on {\em Big} cluster with {\em Kernel-level} split among all four {\em Big} cores and rest $(W-X)$ layers are processed on {\em Small} cluster. 
The challenge is to find an optimal split point \textit{X} with maximum throughput. 
There are ${W-1 \choose 1 }= (W-1)$ possible split points in this pipeline. 
Figure~\ref{fig:2_stage_1_frame_Split} shows throughput for different CNNs with split ratio ({\em X/W}) ranging from zero to one. 
We also include fully-connected layers for \textit{AlexNet} as valid points to split. 
Optimal split ranges from 0.60 for {\em GoogLeNet} to 0.90 for {\em AlexNet}.
\begin{figure}[!t]
	\centering
	\scriptsize
	\begin{tikzpicture}
	\begin{axis}[
	width=\columnwidth, height=\chartheight,
	xlabel={Ratio of Convolutional Layers Split on \textit{Cortex A73} cluster},
	ylabel={Normalized Throughput [Img/s]},
	legend style={at={(0.5,1.05)}, anchor=south, cells={anchor=west}, font=\small, },
	legend columns=5,
	ymajorgrids=true,
	]
	
	\addlegendentry{{\em AlexNet}}
	\addplot[
	only marks, mark=otimes, mark size =  1,
	]
	table[x=Split,y=Throughput]
	{Data/2_stage_alexnet.data};
	
	\addlegendentry{{\em GoogLeNet}}
	\addplot[
	only marks, mark=square, mark size =  1, red,
	]
	table[x=Split,y=Throughput]
	{Data/2_stage_googlenet.data};
	
	\addlegendentry{{\em MobileNet}}
	\addplot[
	only marks, mark=triangle, mark size =  1, blue,
	]
	table[x=Split,y=Throughput]
	{Data/2_stage_mobilenet.data};
	
	\addlegendentry{{\em ResNet50}}
	\addplot[
	only marks, mark=diamond, mark size =  1, lime,
	]
	table[x=Split,y=Throughput]
	{Data/2_stage_resnet50.data};
	
	\addlegendentry{{\em SqueezeNet}}
	\addplot[
	only marks, mark=star, mark size =  1, brown,
	]
	table[x=Split,y=Throughput]
	{Data/2_stage_squeezenet.data};
	
	\end{axis}
	\end{tikzpicture}
	
	\caption{Throughput of a two-stage pipeline {\em (B4-s4)} with workload split at different convolutional layers normalized against the maximum throughput obtained.}
	\label{fig:2_stage_1_frame_Split}
\end{figure}
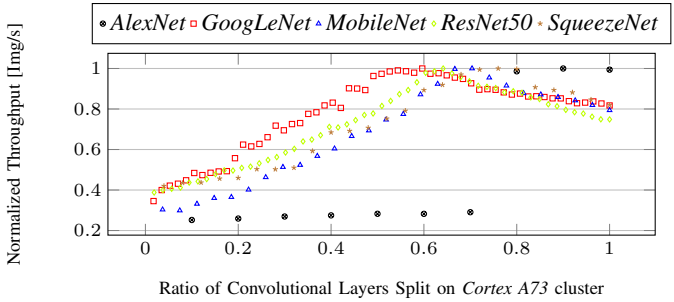
\begin{figure}[tb!]
	\centering
	\begin{tikzpicture}
	\begin{axis} [
	xlabel={Split between s2-s2},
	ylabel={Split between B-s},
	zlabel={Throughput [Img/s]},
	xlabel style={sloped}, ylabel style={sloped},  zlabel style={sloped},
	colorbar horizontal,
	xticklabel={%
		\pgfmathtruncatemacro{\IntegerTick}{\tick}%
		\pgfmathprintnumberto[verbatim,fixed,precision=3]{\tick}\tickAdjusted%
		\pgfmathparse{\IntegerTick == \tickAdjusted ? 1: 0}%
		\ifnum\pgfmathresult>0\relax$\IntegerTick$\else\fi%
	}, 
	yticklabel={%
		\pgfmathtruncatemacro{\IntegerTick}{\tick}%
		\pgfmathprintnumberto[verbatim,fixed,precision=3]{\tick}\tickAdjusted%
		\pgfmathparse{\IntegerTick == \tickAdjusted ? 1: 0}%
		\ifnum\pgfmathresult>0\relax$\IntegerTick$\else\fi%
	},   
	]
	\addplot3+[only marks,mark size=1pt,scatter] table[x={xsplit},y={ysplit},z={value}] {Data/3data.data};
	\end{axis}
	\end{tikzpicture}
	\caption{Throughput of {\em ResNet50} with a three-stage pipeline {\em (B4-s2-s2)} with workload split at different layers. } 
	\label{fig:3_stage_resnet}
\end{figure}
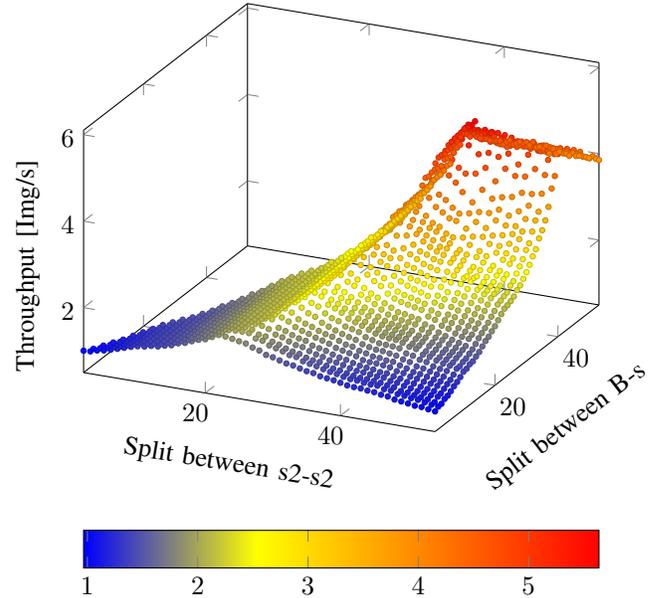

Design space for a three-stage pipeline  is much larger as we need to locate two split points $X_1$ and $X_2$. 
Consider a pipeline configuration {\em (B4-s2-s2)}. Four \textit{Big} cores, two \textit{Small} cores, and remaining two {\em Small} cores are used to construct pipeline Stages~1, 2, and 3, respectively. 
Figure~\ref{fig:3_stage_resnet} shows the execution of \textit{ResNet50} with different configurations. 
The $y$-axis shows split point $X_1$, which splits Stage~1 {\em (B4)} and [Stage~2~+~3] {\em (s2-s2)}. 
$X_1$ also splits {\em Big} and {\em Small} clusters for this pipeline configuration. 
The $x$-axis shows split point $X_2$, which splits Stages 2 {\em (s2)} and 3 {\em (s2)}. 
The $z$-axis shows the throughput for a workload split. 
Throughput peaks at 5.6\,Img/s with split points $X_1$ and $X_2$ at Layers 33 and 45, respectively.
The optimal three-stage pipeline for \textit{ResNet50} has 7\% higher throughput than the corresponding optimal two-stage pipeline.

\subsection{Stages of Pipelines}\label{subsec:pipeline_stage}
We can create pipelines with many more stages (up to $H$ on heterogeneous multi-core with $H$ cores) in pursuit of higher throughput for CNN inference. 
We eliminate pipeline designs with heterogeneous core types within pipeline stage as {\em Kernel-level} split between clusters is not helpful (Figure~\ref{fig:CNN_Default_Split}). 
We only consider  pipeline configurations with {\em Big} cores for initial convolutional layers and {\em Small} cores for subsequent convolutional layers as CNNs usually have more compute-intensive convolutional kernels at the beginning (Figure~\ref{fig:ConvolutionProcessingDistribution}). 

Equation~(\ref{eqn:pipelines}) gives the number of different pipelines possible $C_p$ with $p$ pipeline stages on heterogeneous multi-core with $H_{B}$ {\em Big} cores and $H_{s}$ {\em Small} cores. 
We use $p_B$ and $p_s$ to denote the number of stages constructed with the {\em Big}  and {\em Small} clusters, respectively. 
${H_{B}-1 \choose P_B-1} \times {H_{s}-1 \choose P_s-1}$ gives the total number of different pipeline that we can construct. 
However, the values of $p_B$ and $p_s$ must satisfy the following requirements to construct a meaningful $p$-stage pipeline.
 
\begin{equation*}
p_B \in [1, H_B], \text{   } p_s \in [1, H_s], \text{   } p_B + p_s = p	
\end{equation*}

Thus, the minimum value of $\max(1, p-H_s)$ and the maximum value of $\min(H_B, p-1)$ gives a range of $p_B$. 
We then go through $p_B$ and calculate the total number of different pipelines possible with $p$ stages using Equation~(\ref{eqn:pipelines}).

\begin{equation}\label{eqn:pipelines}\small
C_p = \sum_{P_B=max(1, p-H_{s})}^{min(H_{B}, p-1)}  {H_{B}-1 \choose P_B-1} \times {H_{s}-1 \choose (p-P_B)-1}
\end{equation}

Equation~(\ref{eqn:design_points}) gives the total number of design points for CNN with $W$ convolutional layers ($D_W$) in {\em Layer-level} splitting on $H$-core heterogeneous multi-core.

\begin{equation}\label{eqn:design_points}\small
D_W= \sum_{p=2}^{H} {W-1 \choose p-1} \times C_p
\end{equation}

There are in total 64 possible pipelines (with $p = 2$ to $8$)  as calculated with Equation~\eqref{eqn:pipelines} for our prototype board with eight-core heterogeneous multi-core.
Furthermore, there are in total 5,379,616 distinct possible design points for {\em MobileNet} with its 28 convolutional layers as calculated using Equation~\eqref{eqn:design_points}. 
Design space gets even larger for bigger CNNs like {\em GoogLeNet} and {\em ResNet50} with more layers. 
Therefore, it is not possible to explore entire {\em Layer-level} splitting design space using exhaustive search in a reasonable amount of time. 

\subsection{The \textit{Pipe-it} Framework}

We present a two-part \textit{Pipe-it} framework to quickly go through huge design space and locate the best configuration to execute given CNN workload. 
\textit{Pipe-it} first predicts the execution time of all layers on all possible core configuration from static network-layer configuration descriptors~(Section~\ref{model}). \textit{Pipe-it} then goes through design space heuristically using predicted timing information to obtain near-optimal pipeline configuration and corresponding workload allocation~(Section~\ref{framework}).

\section {Layer-wise Performance Estimation} \label {model}

The most time-consuming part of CNNs is the execution of convolutional layers.
Convolutional layers convolve input tensors with filters to generate respective output tensors, feeding into following layers as inputs. With the extensive calculation requirements, hardware-dependent implementation and optimization techniques are applied to accelerate the execution of convolutions. 

GEMM is commonly used to implement convolution executions. 
\textit{ARM-CL} first converts input image tensor and filter into a matrix ({\em Im2col kernel}). 
It then performs GEMM execution and finally transforms the execution results back into output image tensor format ({\em Col2Im kernel}). 
Authors in~\cite{Lu2017} show execution time of convolution correlates linearly to the dimension of matrices. 
We build on the approach that correlates statically available descriptors of each convolutional layers with layer execution times.
We evaluate and model individual convolutional layers with special consideration on the effects of multi-threading, whereas~\cite{Lu2017} only considers the overall execution time of the network.

\subsection{Convolution as GEMM}

\begin{figure}[!t]
	\centering
	\includegraphics[width=\columnwidth]{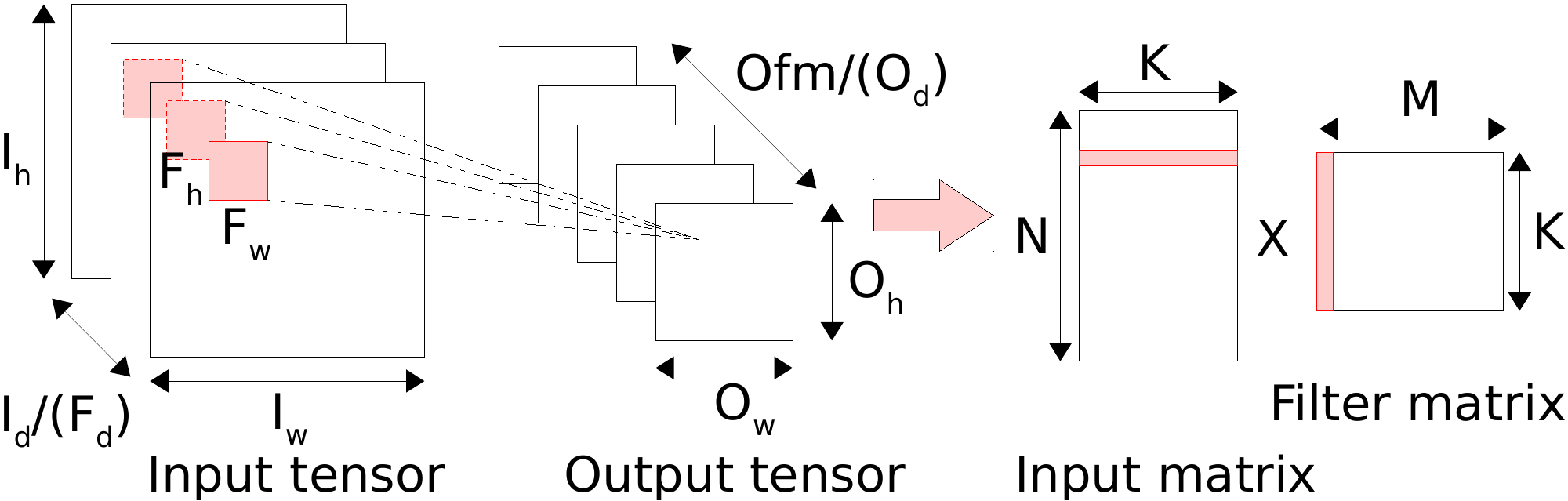}
	\caption{Visualization of a convolutional layer with input image tensor of size $\{ I_w, I_h, I_d \}$ and filter of size $\{ F_w, F_h, F_d, Ofm\}$ generating output tensor of size $\{ O_w, O_h, O_d \}$. 
	The execution is realized as GEMM of input matrix $ [ N \times K] $ and filter matrix $ [K \times M]$ generates result matrix of size $ [N \times M]$.}
	\label{fig:cnn}
\end{figure}

Figure~\ref{fig:cnn} visualizes convolution using GEMM. 
Consider a convolutional layer with input image tensor of size (height, width, depth) $\{ I_w, I_h, I_d \}$ and filter of size (height, width, depth, number of output feature maps) $\{ F_w, F_h, F_d, Ofm\}$, with padding $Pad$, and stride $S$. 
Convolutional layer generates output tensor $\{ O_w, O_h, O_d \}$ of size given by Equation \eqref{eqn:conv_size}. 
Input tensor and filter are required to have matching depth ($I_d = F_d$) and are usually square ($I_w = I_d, O_w = O_h$).
\begin{equation} \label{eqn:conv_size}
\begin{split}
O_w &= \lfloor(I_w -F_w + 2*Pad)/S\rfloor +  1 \\
O_h &= \lfloor(I_h - F_h + 2*Pad)/S\rfloor + 1 \\
O_d &= Ofm
\end{split}
\end{equation}

{\em ARM-CL} implements convolution as GEMM of input and filter matrices. 
Figure~\ref{fig:cnn} shows how the input tensor are divided into small patches of size of one filter ($\{ F_w, F_h, F_d\}$), denoted as the red shaded region. The patches are re-arranged as rows in the image matrix. Similarly, the filters are re-arranged into columns in the filter matrix. Thus the convolution is transformed into a GEMM of an image matrix ($ [ N \times K] $) and a filter matrix ($ [K \times M]$), which generates a result matrix of size $ [N \times M] $ and later resize it into an output tensor. 
Equation~\eqref{eqn:gemm_size} gives dimensions of matrices. 
The total number of arithmetic operations is $( N \times K \times M)$. 
\begin{equation} \label{eqn:gemm_size}
\begin{split}
N = O_w \times O_h \\ 
K = F_w \times F_h \times F_d \\
M = Ofm
\end{split}
\end{equation}

Compute time of GEMM is a complex function of memory accesses, arithmetic computations, and inherent exploitable parallelism in the given convolutional kernel. 

\subsection{Single Core Estimation}

We create a set of micro-benchmarks with {\em ARM-CL} to capture the execution behaviour of layers commonly used in networks. The micro-benchmarks contain representative layers and a convolutional layer with desired configurations (input sizes and filter sizes). We randomly generate input images and filter parameters for measurement purposes. The GEMM execution time is measured for different configuration points using the following values of the parameters:
$$
I_w = I_h = \{ 7, 14, 28, 56, 112 \} $$$$
F_w = F_h = \{ 1, 3, 5, 7, 11 \}$$$$
I_d = F_d = \{ 32, 64, 92, 128, 192, 256 \}$$$$
Ofm = \{ 32, 64, 92, 128, 192, 256 \}$$

We observe a linear correlation between the dimensions of matrices ($ N, K, M $) and the execution time of GEMM. 
Authors in~\cite{Lu2017} made similar observations. 
Equation~\eqref{eqn:regression} models the execution time of convolutional layer $T$  by using linear regression on ($N, K, M$) for a single-core configuration, where $\beta_1$, ($\beta_2$, ..., $\beta_8$) are constants determined with the help of linear regression.
We can physically interpret interaction terms in Equation~\eqref{eqn:regression} as the size of matrices involved in GEMM ($NK, KM, NM $) and total arithmetic operations ($NMK$).
\begin{equation} \label{eqn:regression}
\begin{split}
T = &\beta_1 N + \beta_2 K + \beta_3 M +  \beta_4 NK + \beta_5 KM \\&+ \beta_6 NM + \beta_7 NMK + \beta_8
\end{split}
\end{equation}

\subsection{Multi-core Estimation}

\begin{figure}[!t]
	\centering
	\includegraphics[width=0.8\columnwidth]{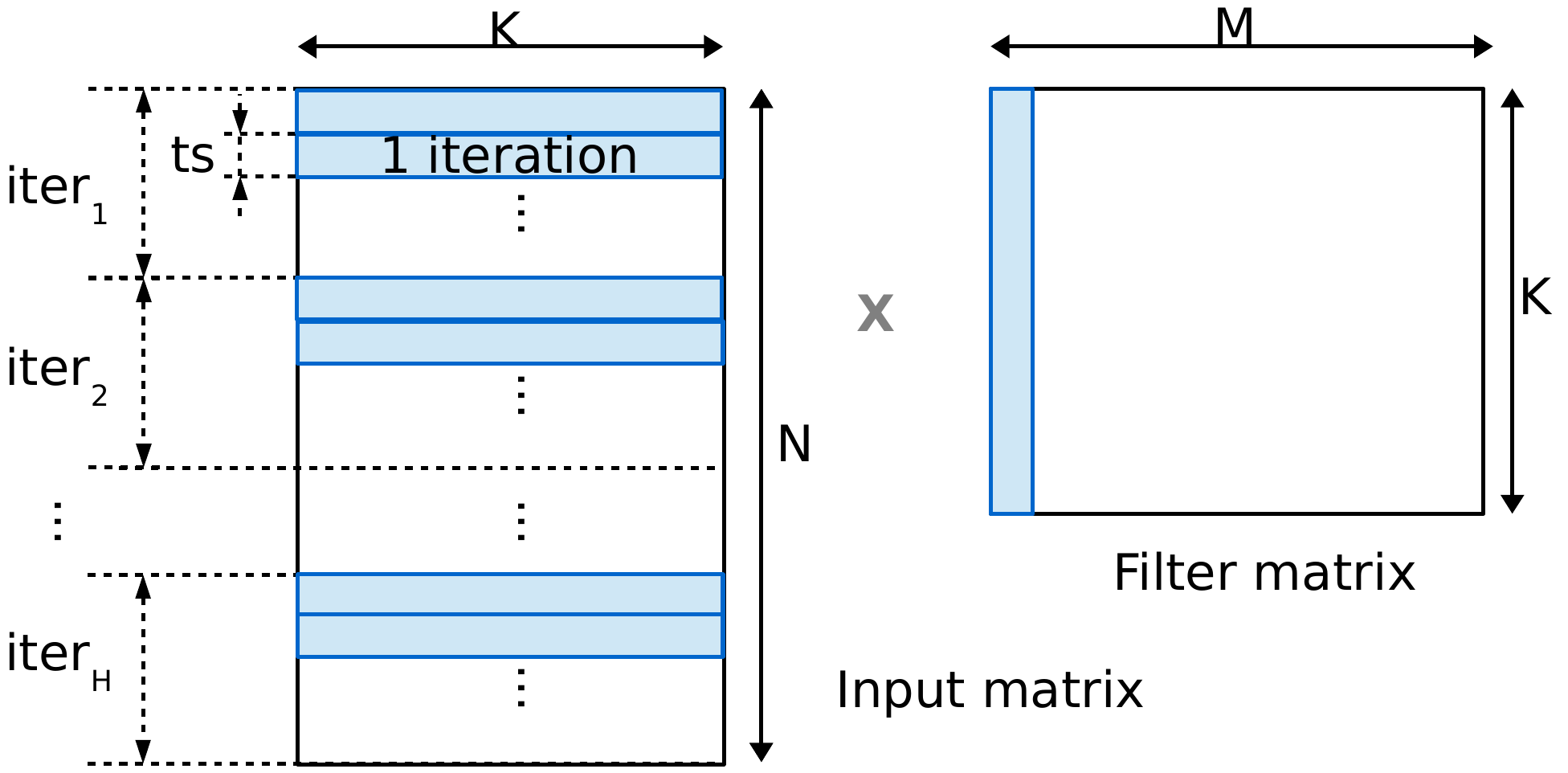}
	\caption{Visualization of iteration allocation for convolutional layer among $H$ threads.}
	\label{fig:iter}
\end{figure}

{\em ARM-CL} implements GEMM optimization by multi-threading and tiling with tile size ($ts$) determined according to the cache sizes to achieve optimal memory behaviour. It uses $H$ threads for execution on an $H$-core multi-core. As shown in Figure~\ref{fig:iter}, the total workload is divided along the rows of the image matrix into chunks of ``iterations''. The total count of iterations is $n_{iter} = N/ts$. These iterations are then dispatched either statically or dynamically to available threads. 
A thread $t$ is assigned with $iter_t$ number of iterations to execute sequentially. 
Workload assigned to all $H$ threads add up to the total number of iterations ($\sum_{t = 1}^{H} iter_t = n_{iter}$).

For single-threaded execution, all iterations ($n_{iter}$) are assigned and processed sequentially on one thread, with execution time $T$ obtained from Equation~\eqref{eqn:regression}. We model the time of each iteration ($T_{iter}$) from the single-threaded execution time with Equation~\eqref{eqn:iteration_time}, assuming identical processing time for all iterations. For multi-thread execution, the execution time of the slowest thread determines the total time when we distribute the workload among $H$ threads, as shown in Equation~\eqref{eqn:multi_time} which models $T_{multi}$. Constant coefficients ($\alpha_1$, $\alpha_2$, $\alpha_3$) are obtained using linear regression.
\begin{equation}\label{eqn:iteration_time}
T_{iter} = (T - \alpha_1 )/ n_{iter}+ \alpha_2
\end{equation}
\begin{equation}\label{eqn:multi_time}
T_{multi} = \max_{t \in [1, H]}( T_{iter} * iter_t )+ \alpha_3
\end{equation}

We can expect an equal split ($iter_t = n_{iter} / H = N/(ts*H)$)  on the distribution of workload among homogeneous cores. 
Equation~\eqref{eqn:multi_time_2} combines the previous two equations and models the multi-threaded execution time $T_{multi}$ based on matrix size $N$, tile size $ts$, and the number of cores $H$. 
\begin{equation}\label{eqn:multi_time_2}
\begin{split}
T_{multi} &= T_{iter} * iter_t + \alpha_3 = T_{iter} * n_{iter} / H + \alpha_3 \\
&= (T - \alpha_1 )/H + \alpha_2 * N / (ts * H) + \alpha_3\\
\end{split}
\end{equation}

Table~\ref{tab:layer_model} shows the prediction error for all the possible homogeneous core allocations. The proposed model predicts execution time for individual convolutional layers across all core configurations for all five benchmark CNNs accurately. 
We observed 13.2\% and 11.4\% prediction errors overall on average for {\em Big} and {\em Small} cores, respectively. Our proposed performance-prediction model is significantly more advanced than the model presented in~\cite{Lu2017} focusing on performance prediction for the entire neural network. Model in~\cite{Lu2017} does not take into consideration the different number of cores involved in CNN execution. Their model is built and tested by profiling on all the available cores. For heavy layers, the workload is more likely to occupy all the cores and thus can be predicted with higher precision by~\cite{Lu2017}. However, for light layers, such a method cannot predict the reduction in utilization and thus results in higher errors. The authors in~\cite{Lu2017} only evaluate the model on the entire network and do not include layer-wise evaluations. They report 13.4\% prediction error for overall CNN inference time with only two CNNs. 
We re-implement the model in~\cite{Lu2017} on four {\em Big} cores and observe on average 15\% estimation error for entire networks across five CNNs. 
However, we observe an average 54\% error when using the same model to predict the execution time of a single layer.
A huge error in layer-wise predictions makes the model in~\cite{Lu2017} unusable for \textit{Pipe-it} that requires accurate per-layer performance estimation for workload allocation.

\subsection{Fully-connected Layers}
We also consider fully-connected layers apart from convolutional layers as major layers in \textit{Pipe-it}. Figure~\ref{fig:Layer_Split} shows older networks like \textit{AlexNet} spend a significant portion of their execution time executing fully-connected layers.
However, fully-connected layers involve a huge number of parameters and hence result in excessive memory transfers during execution~\cite{lin2013network}.
Newer CNNs usually adopt structures with no fully-connected layers (\textit{SqueezeNet}) or only one as classifier at the end (\textit{GoogLeNet}, \textit{MobileNet}, \textit{ResNet50}).  

Fully-connected layers are matrix multiplications. \textit{AlexNet} has three fully-connected layers with 4096, 4096, and 1000 neurons respectively. Other networks employ fully-connected layers as a classifier with 1000 neurons. We generate a set of micro-benchmarks with various input tensor sizes and number of neurons (4096 and 1000). Simple linearity is observed between input tensor sizes and execution time for a given number of neurons. Therefore, the regression-based model can also be used to predict the execution time of fully-connected layers. 
We observed 11.8\% and 14.4\% prediction errors overall on average for the fully-connected layers in micro-benchmarks and actual CNNs, respectively.

\begin{table}[!t]
	\centering
	\caption{GEMM execution time prediction error averaged across all convolutional layers in CNN for different possible homogeneous core allocations. }
	\resizebox{\columnwidth}{!}{
		\begin{tabular}{l|r|r|r|r|r|r|r|r}
			\hline
			\multicolumn{1}{c|}{CNN} & \multicolumn{1}{c|}{1B} & \multicolumn{1}{c|}{2B} & \multicolumn{1}{c|}{3B} & \multicolumn{1}{c|}{4B} & \multicolumn{1}{c|}{1s} & \multicolumn{1}{c|}{2s} & \multicolumn{1}{c|}{3s} & \multicolumn{1}{c}{4s} \\ \hline\hline
			{ AlexNet} & 11.3 & 11.9 & 12.3 & 13.1 & 9.6 & 10.5 & 10.5 & 11.1 \\ \hline
			{ GoogLeNet} & 13.8 & 15.0 & 15.1 & 15.0 & 8.8 & 9.5 & 9.6 & 8.9 \\ \hline
			{ MobileNet} & 21.5 & 19.5 & 17.2 & 17.7 & 18.6 & 17.1 & 17.2 & 18.5 \\ \hline
			{ ResNet50} & 8.2 & 7.5 & 8.0 & 8.4 & 11.5 & 10.9 & 11.1 & 12.1 \\ \hline
			{ SqueezeNet} & 18.1 & 17.9 & 18.0 & 17.7 & 13.9 & 13.0 & 11.8 & 12.7 \\ \hline
			Average & \multicolumn{4}{r|}{13.2\%} & \multicolumn{4}{r}{11.4\%} \\ \hline
		\end{tabular}
	}\label{tab:layer_model}
\end{table}

\section {Design Space Exploration} \label {framework}
We can design many different pipelines with a different number of stages and each stage with different processing core combinations for a heterogeneous multi-core. In addition, for fixed pipeline design, the number of design points in allocating the workload to different pipeline stages grows exponentially with the total number of convolutional layers. 
Therefore, we propose a robust heuristic approach that quickly navigates through the design space to obtain a high-performing layer-level split design point for any CNN. 
The heuristic uses an iterative two-step approach.
The first step is to determine a workload split for a given pipeline configuration~(Section~\ref{work_flow}).
The second step is to merge adjacent stages to search for better pipeline configuration~(Section~\ref{stage_merge}). 
Two steps are iteratively engaged to approach a high-throughout pipeline configuration and corresponding workload distribution.

\subsection{Definitions}

Consider CNN with $W$ convolutional layers to be deployed on $(H_B + H_s)$ heterogeneous multi-core with $H_B$ {\em Big} cores and $H_s$ {\em Small} cores. 
The goal of \textit{Pipe-it} is to find throughput maximizing pipeline configuration $P$ and corresponding layer distribution $L$.

We use $P = \{P_1, ... P_p\}$ to define core configuration of each pipeline stage for a pipeline $P$ with $p$ stages. 
We define the pipeline stage as tuple $P_i = ( core\_type, core\_count)$ depicting type and count of cores that are used to construct it. 
The $core\_type$ can only be either $B$ or $s$ since only homogeneous cores are used to construct the pipeline stage. 
There are $H_B$ and $H_s$ core combinations for {\em Big} and {\em Small} cores, respectively. 
Therefore, $(H_B + H_s)$ different pipeline stage configurations are possible.

$L = \{L_1,..., L_p\}$ defines  corresponding layer allocation associated with the pipeline, where $L_i$ is a set of layers allocated to pipeline stage $P_i$. 
$L_i = \{l_1, ..., l_W\}$ if \textit{Pipe-it}  allocates all the $W$ layers to $P_i$.
$L_i = \emptyset$ if it allocates none of the layers to $P_i$.

Section~\ref{model} describes performance-prediction models used to predict the execution time of a layer.
We use time matrix $T$ to represent predicted execution times. 
$T_{l_j}^{P_i}$ represents execution time for layer  $l_j$ on a core configuration $P_i$.
Similarly, the following equation represents the execution time of the pipeline stage $P_i$ with layer allocation $L_i$.

\begin{equation}
T_{L_i}^{P_i} = \sum_{l_j \in L_i}^{} T_{l_j}^{P_i}
\end{equation}

\subsection{Work-Flow Split Determination}\label{work_flow}
We work with an assumption based on Figure~\ref{fig:ConvolutionProcessingDistribution} that initial CNN layers are more compute-intensive than deeper layers and thereby requires more processing power. Thus, we order the pipeline stages to have more compute capable core combinations at the beginning, and with decreasing compute capability for stages deeper into the pipeline. Such an arrangement also ensures a monotonous increase in layer processing time as we move down pipeline stages. The compute capability of core combinations is evaluated by the execution time of layers on average. Equation~\eqref{eqn:compute_cap} gives observed compute capability in executing layer $l$ with homogeneous core combinations on our heterogeneous eight-core platform.

\begin{equation}\label{eqn:compute_cap}
\begin{split}
T_{l}^{(B,4)} < T_{l}^{(B,3)} < T_{l}^{{B,2}} \lessapprox T_{l}^{(s,4)} \\< T_{l}^{(s,3)} < T_{l}^{(s,2)} \lessapprox T_{l}^{(B,1)} < T_{l}^{(s,1)}
\end{split}
\end{equation}

Equation~\eqref{eqn:tp} gives the throughput of pipeline $P$ with $p$ stages and layer allocation $L$. The pipeline stage that produces the longest latency determines the throughput of the pipeline. Therefore, the goal is to balance the workload among all stages to achieve minimal latency (maximum throughput). 

\begin{equation}\label{eqn:tp}
Throughput = 1 / \max_{i \in [1, p]} (T_{L_i}^{P_i})
\end{equation}

Algorithm~\ref{algo:split} describes the division and allocation of a set of layers $L_{wl} = \{l_{a}, ... , l_{b}\}$ (in the original order) among two adjacent pipeline stages $P_i$ and $P_{i+1}$.
The ordering of pipeline stages ensures that any layer $l_j$  is executed faster on $P_i$ than on $P_{i+1}$ ($T^{P_i}_{l_j} < T^{P_{i+1}}_{l_j}$). 
Such arrangement results in an expansion in execution time as we move deeper into the pipeline and thereby ensures one-way flow of workload. 

The workload initially is entirely allocated to fastest stage $P_i$ ($L_i = \{l_{a}, ... , l_{b}\}, L_{i+1} = \emptyset$) making it the bottleneck. We try to move layers to $P_{i+1}$ to balance workload in each stage, starting with the last layer allocated to $P_i$ (layer $l_b$).
Moving layer $l_{j}$ to $P_{i+1}$ is helpful if ($ T^{P_i}_{L_i} - T^{P_i}_{l_j} > T^{P_{i+1}}_{L_{i+1}} + T^{P_{i+1}}_{l_j}$).
We keep moving the layers until $l_{k}$ when $P_{i+1}$ becomes bottleneck instead. 
Moving of more layers to stage $P_{i+1}$ will make it even slower. 
Thus, the best split between two adjacent pipeline stage will be $L_i= \{l_{a}, ... , l_{k}\}$ and $L_{i+1}= \{l_{k+1}, ... , l_{b}\}$.

\textit{Pipe-it} then goes to the next adjacent pipeline stages ($P_{i+1}$ and $P_{i+2}$) to continue balancing stage latency. 
\textit{Pipe-it} uses Algorithm~\ref{algo:split} to go through all stages in pipeline to balance workload with its immediate next stage. 
We symbolize workload as water that flows from the first pipeline stage to deeper stages. 
There will be more space available in an initial stage once a part of workload flows from it to deeper stages.
Therefore, \textit{Pipe-it} engages Algorithm~\ref{algo:split} iteratively to reach the final splitting configuration, wherein there is no further workload redistribution possible.

\begin{algorithm}[!t]
	\caption{ \textbf{find\_split}: Algorithm to split the workload between adjacent pipeline stages.}
	\begin{algorithmic}[1] \label{algo:split}
		\renewcommand{\algorithmicrequire}{\textbf{Input:}}
		\renewcommand{\algorithmicensure}{\textbf{Output:}}
		\REQUIRE $L_{wl} = \{l_a, ..., l_b\}, T^{P_i}, T^{P_{i+1}}$,
		\ENSURE  $L_i, L_{i+1}$
		\\ \textit{Initialisation} : $L_i = L_{wl} = \{l_a, ..., l_b\}; L_{i+1} = \emptyset$;
		\FOR {$l_j \in L_{wl}$}
		\STATE $T^{P_i}_{new} = T^{P_i}_{L_i} - T^{P_i}_{l_j}$; 
		\STATE $T^{P_{i+1}}_{new} =T^{P_{i+1}}_{L_{i+1}} + T^{P_{i+1}}_{l_j} $;
		\IF {($ T^{P_i}_{new} > T^{P_{i+1}}_{new}  $)}
		\STATE $L_i = L_i \setminus \{l_j\}; L_{i+1} = L_{i+1} \cup \{l_j\}$  //  move of $l_j$ is helpful
		\ELSE 
		\STATE break; //further flow of workload will not be helpful
		\ENDIF
		\ENDFOR
		\RETURN $L_i, L_{i+1}$
	\end{algorithmic}
\end{algorithm}

\begin{algorithm}[!t]
	\caption{ \textbf{work\_flow}: Algorithm for workload allocation for a multi-stage pipeline.}
	\begin{algorithmic}[1] \label{algo:work_flow}
		\renewcommand{\algorithmicrequire}{\textbf{Input:}}
		\renewcommand{\algorithmicensure}{\textbf{Output:}}
		\REQUIRE $P = \{ P_1, ... , P_{p} \}, L_{wl} = \{ l_1, ... , l_{W}\}, T = \{ T^{P_1}, ... ,T^{P_{p}} \},$
		\ENSURE $L$
		\\ \textit{Initialisation} : $L = \{L_1, ..., L_p\}$; \textbf{for}$(L_i \in L)$ \textbf{do} $L_i = \emptyset;$ \textbf{end for}
		\\\hspace{0.6in}$L_1 = L_{wl}; L_{old} = \emptyset;$
		\\ \textit{LOOP: Exit when allocation stabilized}
		\WHILE{ $L \neq L_{old}$}
		\STATE $L_{old} = L$
		\FOR {$P_i, P_{i+1} \in P$}
		\STATE $L_{temp} = L_i \cup L_{i+1}$
		\STATE $L_i, L_{i+1} = $ \textbf{find\_split($L_{temp}, T^{P_i}, T^{P_{i+1}}$)}
		\ENDFOR
		\ENDWHILE
		\RETURN $L$
	\end{algorithmic}
\end{algorithm}

\subsection{Pipeline Stage Merging}\label{stage_merge}

\begin{figure} [!t]
	\centering
	\scriptsize
	
	\begin{tikzpicture}
	\pgfplotsset{set layers}
	\begin{axis}[
	width=\columnwidth, height=\chartheight,
	xlabel={(a) {\em Big} Core Configurations },
	ylabel={Speedup},
	xtick=data,
	xticklabels from table={Data/Core_Concavity_2.data}{Configuration},
	xlabel style={yshift=4pt},
	bar width=3pt,
	ymajorgrids=true,
	yminorgrids=true,
	ymin=1,		
	legend columns = 3,	
	legend style={at={(0.5,1.05)}, anchor=south, font=\scriptsize},
	cycle list name=black white,
	]
	
	\addlegendentry{\em CONV-1}  \addplot [black, mark=otimes]table[x expr = \coordindex, y = CONV-1, ]{Data/Core_Concavity_2.data};
	\addlegendentry{\em CONV-2}  \addplot [red, mark=square] table[x expr = \coordindex, y = CONV-2, ]{Data/Core_Concavity_2.data};		        
	\addlegendentry{\em CONV-3}  \addplot [blue, mark=triangle] table[x expr = \coordindex, y = CONV-3, ]{Data/Core_Concavity_2.data};
	\addlegendentry{\em CONV-4}  \addplot [lime, mark=diamond] table[x expr = \coordindex, y = CONV-4, ]{Data/Core_Concavity_2.data};
	\addlegendentry{\em CONV-5}  \addplot [brown, mark=star] table[x expr = \coordindex, y = CONV-5, ]{Data/Core_Concavity_2.data};
	
	\end{axis}
	\end{tikzpicture}
	
	\begin{tikzpicture}
	\pgfplotsset{set layers}
	\begin{axis}[
	width=\columnwidth, height=\chartheight,
	xlabel={(b) {\em Small} Core Configurations},
	ylabel={Speedup},
	xtick=data,
	xticklabels from table={Data/Core_Concavity.data}{Configuration},
	xlabel style={yshift=2pt},
	bar width=4pt,
	ymajorgrids=true,
	yminorgrids=true,
	ymin=1,		
	cycle list name=black white,
	]
	\addplot [black, mark=otimes]table[x expr = \coordindex, y = CONV-1, ]{Data/Core_Concavity.data};
	\addplot [red, mark=square] table[x expr = \coordindex, y = CONV-2, ]{Data/Core_Concavity.data};		        
	\addplot [blue, mark=triangle] table[x expr = \coordindex, y = CONV-3, ]{Data/Core_Concavity.data};
	\addplot [lime, mark=diamond] table[x expr = \coordindex, y = CONV-4, ]{Data/Core_Concavity.data};
	\addplot [brown, mark=star] table[x expr = \coordindex, y = CONV-5, ]{Data/Core_Concavity.data};
	
	\end{axis}
	\end{tikzpicture}
	
	\caption {The concavity in speedup for the five convolutional layers in {\em AlexNet} with different core configurations.}
	\label{fig:concavity}
\end{figure}
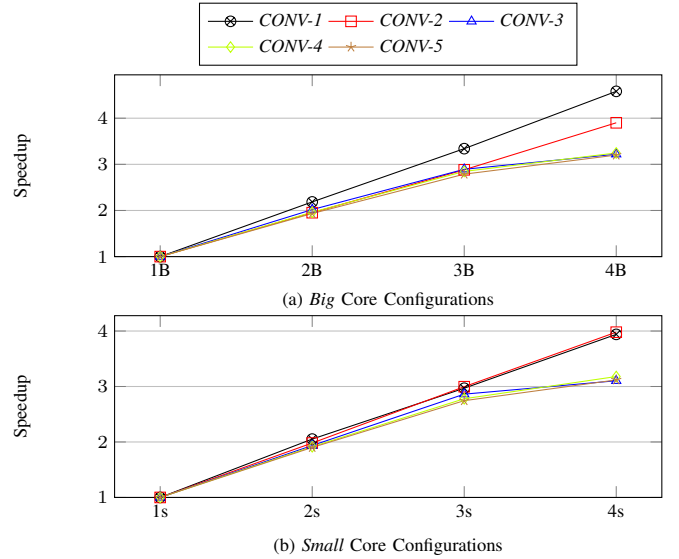

Running GEMM using multi-threading is always beneficial. 
However, Figure~\ref{fig:concavity} shows saturating Thread Level Parallelism~(TLP) can lead to concavity in multi-threaded speedup gains with increasing core allocation.
Furthermore, different types of layers derive different levels of benefits from multi-threading. 
Therefore, it is important to match the size of the pipeline stage with speedup characteristics of layers allocated to it. 
Algorithm~\ref{algo:merge} describes the process of merging pipeline stages to create bigger pipeline stages. 
We consider  {\em Big} cluster first before moving on to {\em Small} cluster.

We start with ($H_B + H_s$)-stage pipeline for ($H_B + H_s$)-core heterogeneous multi-core, where each stage comprises of only one core. 
{\em Pipe-it} engages Algorithm~\ref{algo:work_flow} to search for the best split of workload for this pipeline configuration. 
The pipeline is likely to be bottlenecked by layers that require more compute capability given sub-optimality of single-core performance. 
Thus, we merge pipeline stages to create a more compute capable stage to alleviate the bottleneck. 

Consider the merger of stages $P_i = (core\_type, count_i)$ and $P_{i+1} = (core\_type, count_{i+1})$ to stage $P_{i'} = (core\_type, count_i + count_{i+1})$ with originally allocated set of layers $L_i$ and $L_{i+1}$, respectively.
Note that $P_i$ and $P_{i+1}$ must be of the same core type to merge.

The merging is only helpful when Equation~\eqref{eqn:merge_condition} holds, which implies new stage should be better in performance than at least one of two stages combined. 
Otherwise, we can stop as the concavity in speedup~(Figure~\ref{fig:concavity}) dictates no further merging of the involved stages to create an even bigger stage will be helpful either.

\begin{equation}\label{eqn:merge_condition}
T^{P_{i'}}_{L_{i'}} = T^{P_{i'}}_{L_i} + T^{P_{i'}}_{L_{i+1}} < \max(T^{P_i}_{L_i} , T^{P_{i+1}}_{L_{i+1}})
\end{equation}

Successful merge updates the pipeline configuration and reengages  Algorithm~\ref{algo:work_flow} to find a new higher-performing layer split.
Merging decision depends largely on layers allocated to the stage as different layers respond differently to different stage configurations. 
Therefore, the reallocation of workload is necessary for presenting the right layer information to the merging algorithm.
Algorithm~\ref{algo:merge} runs iteratively until no further merging of stages is helpful. 

\begin{algorithm}[!t]
	\caption{ \textbf{merge\_stage}: Algorithm for determining stage configuration and corresponding workload allocation.}
	\begin{algorithmic}[1] \label{algo:merge}
		\renewcommand{\algorithmicrequire}{\textbf{Input:}}
		\renewcommand{\algorithmicensure}{\textbf{Output:}}
		\REQUIRE $ L_{wl} = \{ l_1, ... , l_{W}\}, H_B, H_s, T$
		\ENSURE  $P, L$
		\\ \textit{Initialisation} :$p = H_B + H_s; P= \{ P_1, ..., P_p\}; L = \{L_1, ..., L_p\} $; 
		\STATE $L = $\textbf{work\_flow($P, L_{wl}, T$)};
		\\ \textit{LOOP: Big} cluster
		\FOR {$ (P_i, P_{i+1}) $ in $ P$}
		\IF {(\textbf{Equation~\eqref{eqn:merge_condition}})}
		\STATE merge, update $P$; $L= $\textbf{work\_flow($...$)};
		\ELSE
		\STATE break; //stop further merging
		\ENDIF
		\ENDFOR
		\\\textit{LOOP: Small} cluster
		\FOR {$ (P_i, P_{i+1}) $ in $ P$}
		\IF {(\textbf{Equation~\eqref{eqn:merge_condition}})}
		\STATE merge, update $P$; $L= $\textbf{work\_flow($...$)};
		\ELSE
		\STATE break; //stop further merging
		\ENDIF
		\ENDFOR
		\RETURN $P, L$
	\end{algorithmic}
\end{algorithm}

\subsection{An Example}
We illustrate with an example of how antecedent algorithms work to locate optimal pipeline configuration and workload allocation. 
The example considers deployment of \textit{ResNet50} with 54 major layers (Table~\ref{tab:cnn_structure}) on an eight-core heterogeneous multi-core with four {\em Big} and four {\em Small} cores.
We can create eight different pipeline stages with different core combinations for this architecture.
Therefore, eight different sets of layer execution time are predicted to generate time matrix $T$ of size (54,8). 
We plug the following corresponding inputs to Algorithm~\ref{algo:merge}.
 
\begin{equation*}\small
L_{wl} = \{l_1, l_2, ..., l_{54}\}; H_B = 4; H_s = 4;
\end{equation*}

Algorithm~\ref{algo:merge} initializes an eight-stage pipeline, wherein each stage consists of only a single core. 
It then engages Algorithm~\ref{algo:work_flow} to find split for the eight-stage pipeline.

Algorithm~\ref{algo:work_flow} allocates all layers to the first pipeline stage $P_1$ at the beginning. 
It then engages Algorithm~\ref{algo:split} to balance the workload between the first two stages ($P_1$ and $P_2$). 
Layers starting with the last layer allocated to $P_1$ (Layer  $l_{54}$) are moved to stage $P_2$ for processing until two stages are balanced.  
Algorithm~\ref{algo:split} returns $L_{1} = \{l_1, ..., l_{25}\}, L_{2} = \{ l_{26},..., l_{54}\}$. 
We use $l_{1-25}$ as a short-hand notation for $\{l_1, ..., l_{25}\}$. 
Thus, {\em Pipe-it} updates the workload allocation to $L = \{l_{1-25}, l_{26-54}, \emptyset, \emptyset,\emptyset,\emptyset,\emptyset,\emptyset\}$.

Algorithm~\ref{algo:work_flow} then continues to balance workload between $P_2$ and $P_3$. 
Algorithm~\ref{algo:work_flow} repeats the process with the remaining pipeline stages.  
The first iteration returns $L = \{l_{1-25}, l_{26-38}, l_{39-46}, l_{47-50}, l_{51}, l_{52-54}, \emptyset, \emptyset\}$. 
The algorithm returns to rebalance workload of $P_1$ and $P_2$ again once it has rebalanced $P_2$ with $P_3$ and other stages. 
The iterative rebalancing, in the end, returns $L = \{l_{1-18}, l_{19-32}, l_{33-41}, l_{42-48}, l_{49-51}, l_{52-54}, \emptyset, \emptyset\}$. 
The last two pipeline stages are not allocated any workload because of poor computation  capabilities. 
Therefore, the merging of stages is necessary to achieve higher performance.

Algorithm~\ref{algo:merge} evaluates a merger of the first two stages $P_1$ and $P_2$ to create a stage comprising of two {\em Big} cores ($(B, 2)$). 
Workload allocation is recalculated with Algorithm~\ref{algo:work_flow} if Equation~\eqref{eqn:merge_condition} holds. 
Otherwise, the merger of stages is not helpful. 
The algorithm will not try with further mergers.
Merger in our case is helpful and algorithm updates the pipeline configuration to $P = \{ (B, 2), (B, 1), (B, 1), (s, 1), (s, 1), (s, 1), (s, 1)\}$, with $L = \{l_{1-29}, l_{30-38}, l_{39-48}, l_{49-51}, l_{52}, l_{53-54}, l_{\emptyset}\}$. 
The algorithm then goes on merging $P_1$ and $P_2$ to create $(B, 3)$ and beyond. 
It recalculates allocation every time the pipeline stage is updated.
The merge goes on for the {\em Small} cluster afterward following similar rules. 
{\em Pipe-it} finally decides upon a three-stage pipeline with configuration $P = \{ (B, 4), (s, 2), (s, 2) \}$ and workload allocation $L = \{ l_{1-35}, l_{36-44}, l_{45-54}\}$.

\section{Experimental Evaluation} \label {Experiments}

\begin{figure}[!t]
	\centering
	\includegraphics[width=\columnwidth]{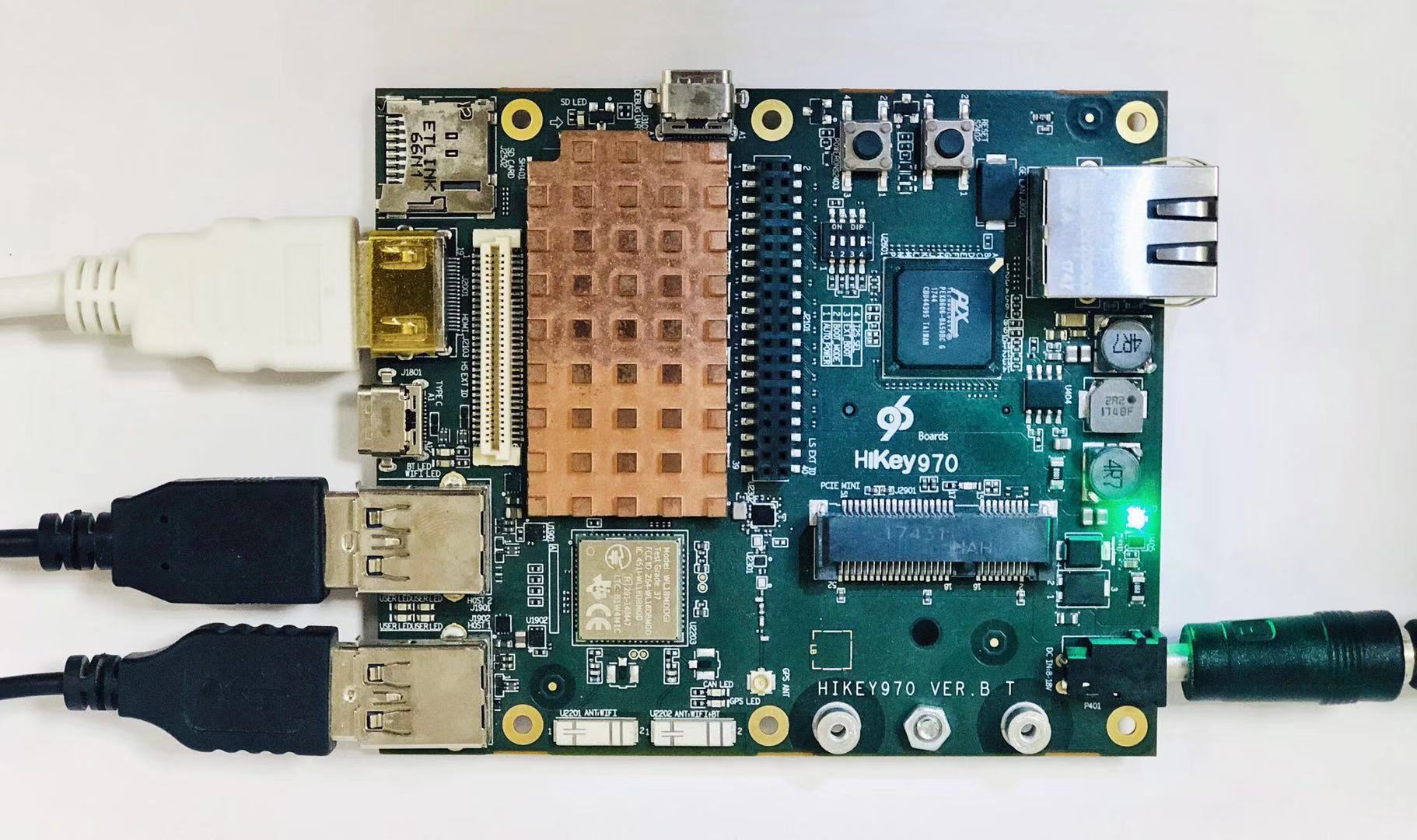}
	\caption{Picture of {\em Hikey 970} mobile development board.}
	\label{fig:hikey970}
\end{figure}

We conduct experimental evaluations on {\em Hikey 970} mobile development platform~\cite{hikey970} for five CNN models as specified in Table~\ref{tab:cnn_structure}. 
Figure~\ref{fig:hikey970} shows a photo of the board in use. 
The board features \textit{ARM big.LITTLE} octa-core CPU with four-core \textit{A73} and four-core \textit{A53} cluster running at the maximum frequency of 2.4\,GHz and 1.8\,GHz, respectively. 
It is connected to a normal desktop monitor through HDMI cable for display. 
It comes equipped with an inbuilt WiFi module through which it can connect with a host machine over Secure SHell (SSH).
Standard DC 5\,V USB fan is used in experiments to eliminate unstable thermal effects.

We classify a continuous stream of 50 images and report average throughput (images processed per second) for each data point. 
The board is left idle for cooling down after each run resulting in approximately a 10\,sec run-time for each point.
An exhaustive search on average size CNN with five million points would take hundred of days to run.
Therefore, the run-time eliminates the possibility of obtaining the optimal configuration using an exhaustive search.

Recall that {\em Kernel-level} split on all eight heterogeneous cores performs worse than four homogeneous {\em Big} cores. 
Therefore, our baseline configuration is {\em Kernel-level} split on four homogeneous {\em Big} cores. 
This baseline provides the best possible throughput with default {\em ARM-CL}~(Table~\ref{tab:algo_res}).

\subsection{Resultant Configurations}
Table~\ref{tab:pipeline_config} shows the outcome of our DSE in the form of pipeline stages $P$ and layer allocation $L$. 
We simplify notation for easier representation. 
For example, the pipeline configuration B4-s2-s2 for {\em ResNet50} implies three pipeline stages consisting of four {\em Big} cores, two {\em Small} cores, and two {\em Small} cores. 
{\em Pipe-it} allocates Layers 1--35, 36--44, and 45--54 to the first, second, and third stage, respectively. 
Table~\ref{tab:algo_res} shows the throughput of the respective pipelines.

In general, throughput benefit of \textit{Pipe-it} comes from a deep and yet balanced pipeline configuration. 
\textit{Pipe-it} can create a better-balanced pipeline with a large number of major layers in a network. 	
Nevertheless, we still observe 20.6\% benefit even for small networks like {\em LeNet} by using a three-stage pipeline designed by~\textit{Pipe-it}, compared to the default execution with 4 cores in the big cluster.
\textit{Pipe-it} on average improves throughput by 39\% over baseline. 
Throughput obtained through pipelined configuration approaches or surpasses combined throughput of individual clusters for all CNNs. 

\begin{table}[!t]
	\centering
	\caption{CNN throughput comparison of homogeneous vs. \textit{Pipe-it} heterogeneous execution with pipelined predicted from actual measured and predicted layer execution time.}
	\resizebox{\columnwidth}{!}{
		\begin{tabular}{l|r|r|r|r|r}
			\hline
			\multicolumn{1}{c|}{\multirow{2}{*}{CNN}} & \multicolumn{2}{c|}{\begin{tabular}[c]{@{}c@{}}Homogeneous \\ Throughput (Imgs/s)\end{tabular}} & \multicolumn{2}{c|}{\begin{tabular}[c]{@{}c@{}}\textit{Pipe-it} -- Heteogeneous \\ Throughput (Imgs/s)\end{tabular}} & \multicolumn{1}{c}{\multirow{2}{*}{\begin{tabular}[c]{@{}c@{}}Percentage\\ Benefit\\ (\%)\end{tabular}}} \\ \cline{2-5}
			\multicolumn{1}{c|}{} & \multicolumn{1}{c|}{\begin{tabular}[c]{@{}c@{}}{\em Big} \\ Cluster\end{tabular}} & \multicolumn{1}{c|}{\begin{tabular}[c]{@{}c@{}}{\em Small} \\ Cluster\end{tabular}} & \multicolumn{1}{c|}{\begin{tabular}[c]{@{}c@{}} with \\ measured \\ layer time\end{tabular}} & \multicolumn{1}{c|}{\begin{tabular}[c]{@{}c@{}} with \\ \textbf{predicted} \\ layer time \end{tabular}} & \multicolumn{1}{c}{} \\ \hline\hline
			AlexNet & 8.1 & 1.5 & 8.9 & \textbf{8.9} & 9.8 \\ \hline
			GoogLeNet & 7.8 & 3.3 & 11.8 & \textbf{11.3} & 45.5 \\ \hline
			MobileNet & 17.4 & 6.6 & 24.0 & \textbf{23.5} & 35.5 \\ \hline
			ResNet50 & 3.1 & 1.5 & 5.5 & \textbf{5.2 } & 67.5 \\ \hline
			SqueezeNet & 15.6 & 6.9 &21.4 & \textbf{21.4} & 37.5 \\ \hline
			\multicolumn{5}{r|}{Average} & 39.2\% \\ \hline
		\end{tabular}	\label{tab:algo_res}
	}
\end{table}

\begin{table}[!t]
	\centering
	\caption{ Best throughput pipeline configuration with \textit{Pipe-it} and respective layer allocations from layer performance-prediction model.}
	\resizebox{\columnwidth}{!}{
		\begin{tabular}{l|l|l}
			\hline
			\multicolumn{1}{c|}{CNN} & \multicolumn{1}{c|}{Pipeline Config.} & \multicolumn{1}{c}{Layer allocation} \\ \hline\hline
			AlexNet & B4 - s4 & {[}1,9{]} - {[}10,11{]} \\ \hline
			GoogLeNet & B4 - s2 - s1 - s1 & {[}1,29{]} - {[}30,41{]} - {[}42,45{]} - {[}46,58{]} \\ \hline
			MobileNet & B2 - B2 - s3 - s1 & {[}1,11{]} - {[}12,21{]} - {[}22,26{]} - {[}27,28{]} \\ \hline
			ResNet50 & B4 - s2 - s2 & {[}1,35{]} - {[}36,44{]} - {[}45,54{]} \\ \hline
			SqueezeNet & B4 - s4 & {[}1,16{]} - {[}17,26{]} \\ \hline
		\end{tabular}
	}\label{tab:pipeline_config}
\end{table}

\begin{table}[!t]
	\centering
	\caption{Best throughput pipeline configuration with \textit{Pipe-it} and respective layer allocations from actual measured layer timings.}
	\resizebox{\columnwidth}{!}{
		\begin{tabular}{l|l|l}
			\hline
			\multicolumn{1}{c|}{CNN} & \multicolumn{1}{c|}{Pipeline Config.} & \multicolumn{1}{c}{Layer Allocation} \\ \hline\hline
			AlexNet & B4 - s4 & {[}1,9{]} - {[}10,11{]}\\ \hline
			GoogLeNet & B4 - s2 - s1 - s1 & {[}1,25{]} - {[}26,39{]} - {[}40,44{]} - {[}45,58{]} \\ \hline
			MobileNet & B2 - B2 - s3 - s1 & {[}1,11{]} - {[}12,19{]} - {[}20,26{]} - {[}27,28{]}\\ \hline
			ResNet50 & B2 - B2 - s3 - s1 & {[}1,16{]} - {[}17,34{]} - {[}35,47{]} - {[}48,54{]}\\ \hline
			SqueezeNet & B4 - s4 & {[}1,19{]} - {[}20,26{]}\\ \hline
		\end{tabular}
	}\label{tab:meaured_config}
\end{table}

\subsection{Layer Performance-Prediction Model}

We use micro-benchmarks to create our layer performance-prediction model.
Predicted layer execution time guide the search for optimal configuration.
Table~\ref{tab:layer_model} shows the model has good accuracy with on average overall prediction error of 13.2\%  and 11.4\% for {\em Big} and {\em Small} clusters, respectively.

Table~\ref{tab:meaured_config} shows {\em Pipe-it} pipeline configurations with actual measure layer timings instead of predicted timings.
The configurations in Table~\ref{tab:pipeline_config} and Table~\ref{tab:meaured_config} are the same in most cases.
There is a mere 4\% difference in performance in the worst-case.
Results establish the efficacy of our model.

\subsection{General Applicability}
\textit{Pipe-it} is applicable across different heterogeneous multi-cores that have at least two clusters. 
We run \textit{Pipe-it}, on the same {\em Hikey 970} platform, but with one {\em Big} core and two {\em Small} cores turned off to simulate an arbitrary {\em big.Little} CPU configuration.
The layer timing estimations obtained as before are plugged into the design space exploration algorithm to locate the best pipeline configuration with the remaining three {\em Big} cores and two {\em Small} cores. Table~\ref{tab:3b2s} shows the results obtained.
\textit{Pipe-it} predicts pipeline configurations with both clusters engaged.
The performance benefit is not as significant compared to results shown in Table~\ref{tab:algo_res} with 4 \textit{Big} and 4 \textit{Small} cores. This is because, in this CPU configuration, only 2 small cores are additionally engaged in the pipeline. Less additional resources result in lower performance improvement.

\begin{table}[!t]
	\centering
	\caption{Benefit of \textit{Pipe-it} on a non-standard configuration with three {\em Big} cores and two {\em Small} cores.}
	\resizebox{\columnwidth}{!}{
	\begin{tabular}{l|r|r|r|r|r}
		\hline
		\multicolumn{1}{c|}{\multirow{2}{*}{CNN}} & \multicolumn{3}{c|}{Throughput (Imgs/s)} & \multicolumn{1}{c|}{\multirow{2}{*}{Config.}} & \multicolumn{1}{c}{\multirow{2}{*}{\begin{tabular}[c]{@{}c@{}}Pct.\\ Benefit (\%)\end{tabular}}} \\ \cline{2-4}
		\multicolumn{1}{c|}{} & \multicolumn{1}{l|}{3 \textit{Big}} & \multicolumn{1}{l|}{2 \textit{Small}} & \multicolumn{1}{l|}{\textit{Pipe-it}} &\multicolumn{1}{c|}{} & \multicolumn{1}{c}{} \\ \hline\hline
		AlexNet & 5.7 & 0.7 & 5.8 & B3-s2 & 1.5 \\ \hline
		GoogLeNet & 6.2 & 0.7 & 7.4 & B3-s2 & 19.3 \\ \hline
		MobileNet & 14.2 & 3.7 & 15.3 & B3-s2 & 7.1 \\ \hline
		ResNet50 & 2.8 & 1.0 & 3.7 & B3-s1-s1 & 31.5 \\ \hline
		SqueezeNet & 11.4 & 3.6 & 13.5 & B3-s2 & 18.1 \\ \hline
		\multicolumn{5}{r|}{Average} & 15.5\% \\ \hline
	\end{tabular}
	}

	\label{tab:3b2s}
\end{table}

\subsection{Power Efficiency}
We are not able to obtain the individual power values of each CPU component due to lack of power sensors in our development board. 
We instead utilize a power measurement module~\cite{pmu} that supplies and measures whole board power consumption. 
Cluster not engaged in execution during homogeneous runs is turned off to eliminate its contribution to total power consumption. 
Measured  whole board socket power $P$ includes everything on board beside CPU.
We mitigate the effect of non-CPU components on total power by subtracting off idle power $P_I$. 
Idle power can vary with several factors. 
Therefore, we measure it again before each run.
The active power readings reported are $P_A = P - P_I$.
Table~\ref{tab:power} shows power measurements and corresponding power-efficiency.

\begin{table}[!t]
	\centering
	\caption{Average power (W) and power-efficiency (Imgs/J) for execution on homogeneous cores and with {\em Pipe-it}.}
	\resizebox{\columnwidth}{!}{
		\begin{tabular}{l|r|r|r|r|r|r}
			\hline
			\multicolumn{1}{c|}{\multirow{2}{*}{CNN}} & \multicolumn{3}{c|}{\begin{tabular}[c]{@{}c@{}}Average Active \\ Power (W)\end{tabular}} & \multicolumn{3}{c}{\begin{tabular}[c]{@{}c@{}} Power Efficiency \\ (Imgs/J)\end{tabular}}\\ \cline{2-7} 
			\multicolumn{1}{r|}{} & {\em Big} & {\em Small} & \textit{Pipe-it} & {\em Big} & {\em Small} & \textit{Pipe-it}  \\ \hline \hline
			AlexNet & 3.8 & 0.7 & 5.1 & 2.1 & 2.1 & 1.8 \\ \hline
			GoogLeNet & 4.6 & 1.1 & 6.6 & 1.7 & 3.1 & 1.7 \\ \hline
			MobileNet & 4.2 & 1.0 & 5.9 & 4.2 & 6.6 & 4.0 \\ \hline
			ResNet50 & 4.0 & 1.0 & 6.5 & 0.8 & 1.5 & 0.8 \\ \hline
			SqueezeNet & 4.9 & 1.3 & 6.9 & 3.2 & 5.5 & 3.1 \\ \hline
		\end{tabular}
	}\label{tab:power}
\end{table}

We cannot separate active memory power from the CPU.
Therefore, power-efficient {\em Small} cluster shows lower than expected power-efficiency for memory intensive CNNs like {\em AlexNet}.  
We attribute the lower power-efficiency with {\em Pipe-it} to extra memory power consumed  due to coherency between different core clusters.

\subsection{Quantization Considerations}

\begin{figure} [!t]
	\centering
	\scriptsize
	\begin{tikzpicture}
	\begin{axis}[
	ybar stacked,
	ymin=7.5,
	ymax=60,
	bar width=15pt,
	width=\columnwidth, height=\chartheight,
	legend columns=3,
	enlargelimits=0.15,
	legend style={at={(0.5,1.05)}, anchor=south, cells={anchor=west}, font=\small, },
	ylabel={Layer Processing Time [ms]},
	symbolic x coords={{\em v18.05-F32}, {\em v18.05-QASYMM8}, {\em v18.11-F32}, {\em v18.11-QASYMM8}},
	xtick=data,
	]
	
	\addplot+[ybar, bar shift=-5, black, fill=white, postaction={ pattern=crosshatch}] plot coordinates {({\em v18.05-F32}, 39.133) ({\em v18.05-QASYMM8}, 33.640) ({\em v18.11-F32}, 33.537) ({\em v18.11-QASYMM8},25.444)};
	
	\addplot+[ybar, bar shift=-5, black, fill=white, postaction={ pattern=north east lines}] plot coordinates {({\em v18.05-F32},13.009) ({\em v18.05-QASYMM8},17.624) ({\em v18.11-F32}, 4.150) ({\em v18.11-QASYMM8},10.891)};
	
	\addplot+[ybar, bar shift=-5, black, fill=white] plot coordinates {({\em v18.05-F32},4.203) ({\em v18.05-QASYMM8},4.932) ({\em v18.11-F32},7.369) ({\em v18.11-QASYMM8},0.186)};
	
	\addplot+[ybar, bar shift=-5, black, fill=white, postaction={ pattern=dots}] plot coordinates {({\em v18.05-F32},0) ({\em v18.05-QASYMM8},0) ({\em v18.11-F32},0) ({\em v18.11-QASYMM8},0)};
	
	\legend{ Convolutional, DW-Convolutional, Others, {\em Pipe-it} effective}
	\end{axis}
	
	\begin{axis}[
	ybar,
	ymin=7.5,
	ymax=60,
	bar width=15pt,
	width=\columnwidth, height=\chartheight,
	enlargelimits=0.15,
	legend style={at={(0.5,1.05)}, anchor=south, cells={anchor=west}, font=\small, },
	ylabel={Layer Processing Time [ms]},
	symbolic x coords={{\em v18.05-F32}, {\em v18.05-QASYMM8}, {\em v18.11-F32}, {\em v18.11-QASYMM8}},
	xtick=data,
	xticklabels={},
	]
	
	\addplot+[ybar, bar shift=10, black, fill=white, postaction={ pattern=dots}] plot coordinates {
		({\em v18.05-F32},43.5)
		({\em v18.05-QASYMM8},45.2)
		({\em v18.11-F32},34.6)
		({\em v18.11-QASYMM8},30.7)};
	\end{axis}
	\end{tikzpicture}

	\caption{Performance comparison of \textit{MobileNet} with quantization across two {\em ARM-CL} versions (v18.05 and v18.11).}
	\label {fig:quantization}
	
\end{figure}
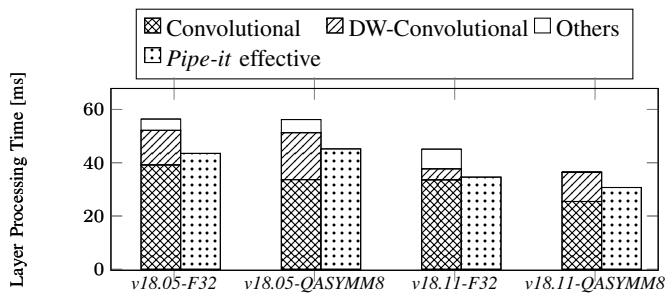

\textit{Pipe-it} aims to improve throughput by engaging all on-chip CPU resources for execution. 
It is orthogonal to optimization techniques such as quantization~\cite{Wu_2016_CVPR}. 
\textit{ARM-CL} provides support for execution with quantized 8-bit using asymmetric integers (QASYMM8). 
However, the benefit of quantization is largely dependent on the implementation. 
Overheads induced by de-quantization and re-quantization operations subdue the benefits of quantization~\cite{sun2017enabling}. 
Figure~\ref{fig:quantization} shows a similar effect by comparing the execution of nonquantized F32 and QASYMM8 for \textit{MobileNet} with \textit{ARM-CL}. 
Execution of convolutional layers is improved by 14\%. However, overall execution time remains unchanged for \textit{ARM-CL v18.05}.

We also evaluate the effect of quantization on the latest \textit{ARM-CL version 18.11}.
F32 implementation of {\em MobileNet} executes 20\% faster on \textit{ARM-CL v18.11} compared to \textit{ARM-CL v18.05}. 
Its convolutional layers are 24\% faster with quantization with an overall 19\% faster execution.

The performance we report above is with homogeneous cores only.
We create pipelines for both original and quantized \textit{MobileNet} using \textit{Pipe-it} across {\em ARM-CL} versions. 
Figure~\ref{fig:quantization} shows the effective per-frame latency (inverse of throughput). 
\textit{Pipe-it} introduces further performance improvement in all implementations. 
\textit{MobileNet} reaches a throughput of 31\,Img/sec with  \textit{Pipe-it} for its quantized \textit{ARM-CL v18.11} implementation.

\subsection{Comparison with Other Frameworks}
We compare the performance of {\em Pipe-it} against other CNN frameworks using {\em MobileNet} as the common denominator.
Figure~\ref{fig:compare_framework} shows the performance of several frameworks.
We measure the Performance of \textit{TVM}, \textit{NCNN}, \textit{Pipe-it}, and \textit{Pipe-it**} with actual experiments on our platform.
Performance numbers for the remaining frameworks are taken from other sources~\cite{ncnn,ai-bench}. 
The borrowed numbers are scaled approximately to compensate for differences in platforms.
\textit{Pipe-it} provides the highest performance amongst all cores.

\begin{figure}[!t]
	\centering
	\scriptsize
	\begin{tikzpicture}
	\begin{axis}[
	ybar,
	bar width=15pt,
	width=\columnwidth, height=\chartheight,
	enlargelimits=0.15,
	legend style={at={(0.5,1.05)}, anchor=south, cells={anchor=west}, font=\small, },
	ylabel={Effective Throughput [Img/s]},
	symbolic x coords={{\em Caffe-android-lib*}, {\em mini-Caffe*}, {\em TF-lite*}, {\em TVM}, {\em NCNN},{\em Pipe-it},{\em Pipe-it**}},
	xtick=data,
	x tick label style={rotate=45,anchor=east},
	]
	\addplot+[ybar, black, fill=white, postaction={ pattern=dots}] plot coordinates {
		({\em Caffe-android-lib*}, 3.7) 
		({\em mini-Caffe*}, 4.3) 
		({\em TF-lite*}, 9.0) 
		({\em TVM}, 5.4) 
		({\em NCNN}, 19.6) 
		({\em Pipe-it}, 23.5) 
		({\em Pipe-it**}, 31)
	};
	\end{axis}
	\end{tikzpicture}
	* scaled performance with \textit{AI-benchmark}~\cite{ai-bench}.\\
	** \textit{Pipe-it} with \textit{ARM-CL v18.11} and quantization as shown in Figure~\ref{fig:quantization}.
	\caption{Performance comparison of \textit{MobileNet} with several frameworks.}
	\label {fig:compare_framework}
\end{figure}

We also compare the energy-efficiency of \textit{Pipe-it} against {\em DeepX}~\cite{Lane2016}. 
{\em DeepX} is designed to consume the least power within a latency requirement. 
Authors of~\cite{Lane2016} evaluate {\em DeepX} on {\em Qualcomm Snapdragon 800} SoC with {\em Krait} four-core 2.3\,GHz CPU. 
{\em DeepX} provides a configuration which consumes 444\,mJ of energy for \textit{AlexNet} with the latency requirement of 500\,ms~(2\,Img/s) resulting in energy-efficiency of 2.2\,Img/J. 
\textit{Pipe-it} achieves comparable energy-efficiency of 1.8\,Img/J but with a much higher throughput of 8.9\,Img/s.

\section {Related Work} \label{Related}
The development of CNNs is moving towards more complex network structures with moderate resource requirements. Starting with 250MB for AlexNet [16] in 2012, the size of models has reduced to less than 0.5MB for SqueezeNet [13] in 2016 without losing accuracy. Such advancements allow for CNN deployment on mobile platforms even with their limited computational and memory resources. To effectively deploy CNN on embedded platforms, researchers are approaching from different angles. The network structure is modified to
fit on the resource-constrained mobile platform, such as quantization~\cite{Wu_2016_CVPR} that accelerates the computation and reduces the memory usage, and network pruning~\cite{yang2018netadapt} that compromise the accuracy with fewer resource requirements.
In addition, sparsity is exploited in NN applications~\cite{sen2018sparce} to reduce the computation and improve execution performance on edge devices.

Accelerators enable highly energy-efficient execution of CNNs on edge devices. 
Several works rely on the computational capability of embedded GPUs to enable CNN with collaborative execution on CPUs and other processors. 
{\em DeepX}~\cite{Lane2016} framework enables NN on edge through co-execution on multiple processors, including GPU and low power processors (LPU). 
It first engages runtime layer compression to control the resource requirement of an NN workload. It then decomposites the workload  into unit-blocks for assignment to multiple processors. {\em DeepX} derives substantial benefit in performance and energy for {\em AlexNet} mainly from its fully-connected layers. Use of fully-connected layers is now minimal in state-of-the-art CNNs. {\em DeepSense}~\cite{huynh2016deepsense} and {\em DeepMon}~\cite{Huynh2017} present an {\em OpenCL} based framework for mobile GPUs. 
{\em DeepSense} adopts GPU memory management techniques which accelerate compute-heavy executions including convolutional and fully-connected layers execution on GPU. {\em DeepMon} extends {\em DeepSense} to include further caching optimizations and improves convolutional layer implementation. ASICs are now being designed specifically for neural network processing such as {\em Google's} Tensor Processing Unit~(TPU) and {\em Huawei's} Neural Processing Units~(NPU). 
Researchers also co-design algorithm and architecture with application-specific characteristics~\cite{zhu2018mobile,zhu2018isca}. 

Researchers have characterized resource requirements of CNNs~\cite{Lane2015,Lu2017} that provide insights on designing CNN with resource-constraints. 
Efficient libraries~\cite{arm_cl,ncnn,chen2018tvm,oskouei2016cnndroid} are created to facilitate the implementation of deep learning on edge devices. 
Frameworks, like {\em CGOOD}~\cite{kang2018cgood} are created to facilitate  deployment of CNN on edge devices by automatically generating {\em C} and GPU ({\em CUDA} or {\em OpenCL}) code  that runs on respective platforms with hardware specifications and optimization requirements.

On the other hand, older technology node or cost-sensitive platforms that lack capable GPUs and accelerators still need to execute CNNs via their CPUs.
{\em Graphi}~\cite{tang2018scheduling} presents a framework that accelerates deep learning models through layer-level parallelism within NN on many-cores. 
It leverages on the inherent layer-level parallelism in network structure and schedules independent layers for concurrent execution. {\em Graphi} is beneficial for networks such as \textit{LSTM} and \textit{GoogLeNet} that have high layer-level parallelism. In comparison, \textit{Pipe-it} looks at computational kernel-level parallelism. It applies to general network structures and targets CNN acceleration on heterogeneous multi-cores.

\section {Conclusion} \label {Conclusion}
On-chip inference using CNNs is now becoming commonplace on edge devices. 
We show in this work that {\em Kernel-level} splitting across heterogeneous core types is detrimental to throughput. Instead, {\em Layer-level} splitting that minimizes cross-cluster coherency can be employed to improve inferencing throughout. 
We introduce a layer-level splitting technique called \textit{Pipe-it} that efficiently uses entire heterogeneous multi-cores to improve CNN inference throughput. 
We study the design space involved and introduce a search algorithm to locate a high performing design point within it. \textit{Pipe-it} improves the throughput on average by 39\% using all heterogeneous cores in comparison to using only homogeneous cores. \textit{Pipe-it} is not limited to CNN applications and also applies to other streaming applications that show similar behaviours. In future, we plan to include more co-processors such as GPUs and NPUs into the design space to further exploit the potential of the embedded SoCs in enabling deep learning.


{
	\bibliographystyle{plain}
	\bibliography{./references/references}  
}

\end{document}